\documentclass[11pt]{article}

\usepackage{array}

\usepackage[preprint]{acl}

\usepackage{times}
\usepackage{latexsym}
\usepackage{amssymb}
\usepackage{amsmath}
\usepackage[T1]{fontenc}

\usepackage[utf8]{inputenc}

\usepackage{microtype}

\usepackage{inconsolata}

\usepackage{graphicx}
\usepackage{booktabs}
\usepackage{arydshln}
\usepackage{multirow}
\usepackage{xspace}
\usepackage{array}
\usepackage{float}
\usepackage{wrapfig}

\title{Predict, Don't Iterate: Efficient Adaptive-Length Infilling\\for Diffusion Language Models}

\author{
\textbf{Haobo Xu$^1$, Sirui Chen$^1$, Yuanchen Bei$^1$, Lingjie Chen$^1$, Yuchen Yan$^1$,}\\ {\textbf{Dongqi Fu$^2$, Jingrui He$^1$, Hanghang Tong$^1$}} \\
$^1$ University of Illinois at Urbana-Champaign\ 
   $^2$ Meta\\
   \texttt{haoboxu@illinois.edu}
  }

\newcommand{\ours}{\textsc{Pill}\xspace}
\newcommand{\pid}[2]{%
  \begin{tabular}[t]{@{}c@{}}
    \texttt{#1}\\[-1.0ex]
    {\scriptsize\texttt{(#2)}}
  \end{tabular}%
}
\begin{document}
\maketitle

\begin{abstract}
Diffusion language models (DLMs) have emerged as a promising alternative to the auto-regressive paradigm. With bidirectional attention and any-order generation, DLMs naturally fit infilling tasks, which require generating a middle span conditioned on both the prefix and the suffix. However, infilling is sensitive to the length of the span, while DLMs require the length to be fixed before generation. Although prior studies extend DLMs to dynamic lengths, they still suffer from two limitations. (i) \textit{Sensitivity to initial length}. These methods require a preset length to initialize the search and are highly sensitive to this initial length, often yielding suboptimal results. (ii) \textit{Inference inefficiency}. They either insert length-changing operations during generation or repeatedly search for an appropriate length using multi-step denoising confidence, both of which introduce substantial extra forward passes and computational cost. Therefore, we propose \ours (\textbf{P}robing-based \textbf{I}nfi\textbf{L}ling with preset-\textbf{L}ength-free decoding), an efficient infilling method for DLMs that requires no preset initial length and adds far fewer extra forward passes than baselines, substantially reducing inference time. Experiments show that, across five DLMs spanning different families, architectures, and training recipes on eight infilling benchmarks, \ours improves over the strongest baseline by $+4.8$ average pass rate on code and $+6.0$ BLEU-2 on text, while running $1.82\times$ faster than that baseline. The code is available at \url{https://github.com/Hsu1023/PILL}.
\end{abstract}
\section{Introduction}
Recently, diffusion language models (DLMs) have emerged as a promising alternative to auto-regressive (AR) paradigm, showing great potential across various tasks~\citep{llada,lladamoe,llada1.5,dream}. Unlike AR models that rely on causal attention and decode strictly left to right, DLMs employ bidirectional attention and generate in any order, allowing them to condition on both the left and right context simultaneously~\citep{hoogeboom2021argmax}. This property makes DLMs particularly well suited to infilling, generating a middle span that is coherent with both a given prefix and a given suffix, a setting that AR models are hard to handle~\citep{humaneval-infilling}.

\begin{figure*}[!t]
    \centering
    \includegraphics[width=1.0\linewidth]{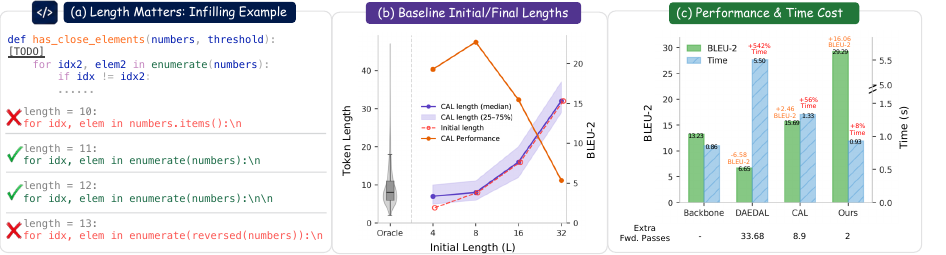}
    \caption{\textbf{Motivation and Performance of \ours.} (a) For DLM infilling tasks, correctness is sensitive to the span lengths. (b) Baselines must preset an initial length $L$ and their final length stays tied to it, whereas \ours requires no preset length. (c) \ours attains the best performance with minimal overhead compared to fixed-length generation with backbones. Baselines are reported at initial length $L=\{4,8,16,32\}$.}
    \label{fig:teaser}
\end{figure*}

However, DLMs require the number of masked positions to be fixed before decoding, whereas the correct infilling length is unknown in advance and varies from case to case~\citep{llada2.0}. 
Crucially, infilling quality is highly sensitive to the chosen length: as shown in Figure~\ref{fig:teaser}(a), the code-infilling example is solved correctly only when the mask span is given exactly the right number of tokens, while slightly shorter or longer spans produce incorrect completions. The model must therefore determine a length before it knows what to generate, even though the appropriate length depends on the content itself. Recent studies tackle this through adaptive-length decoding~\citep{daedal,dreamon,flexmdm,cal}, but they still suffer from two limitations.

First, existing methods are \textbf{\textit{sensitive to the initial length}}. They require a preset length to start the search for a better infilling length. For example, CAL~\citep{cal} will be trapped into local minima near the initial length, yielding varying performances, as shown in Fig.~\ref{fig:teaser}(b). Second, they incur \textit{\textbf{substantial inference overhead}}. Methods either inject explicit one token length-changing operations~\citep{dreamon,daedal}, or repeatedly probe denoising confidence over many decoding steps to estimate a length~\citep{cal}, introducing many extra forward passes and cost.

To this end, we propose \ours, a framework that enables fixed-length DLMs to perform variable-length infilling with only two extra forward passes. \ours consists of three stages. (1) Instead of searching for a length, we \emph{directly predict} it from the bidirectional hidden state of a single mask token inserted between the prefix and suffix. (2) Since generation is sensitive to length, we expand this estimate into a small set of candidate lengths and decode them \emph{in parallel} at a cost close to decoding one. (3) Since DLMs cannot score a generated sequence by next-token likelihood, we \emph{select} the final span with a post-hoc score that favors candidates coherent both internally and with the suffix, using one additional forward pass. In summary, it is more efficient than baselines: it \textit{replaces their iterative length search}, which spends a forward pass per step, with just two extra passes, while the multi-slot decoding generates all candidates in parallel at almost no extra cost. As shown in Figure~\ref{fig:teaser}(c), it achieves the highest quality while adding $<10\%$ wall-clock time over vanilla fixed-length decoding, whereas baselines pay far more for a smaller gain. Our contributions are three-fold:

\begin{itemize}
    \vspace{-0.4em}\item 
\textbf{A preset-free infilling framework}. We propose \ours, a backbone-frozen framework that enables DLMs to perform adaptive-length infilling beyond the fixed lengths with only two extra forward passes, centered on a length probe that predicts target lengths
directly from a mask token's bidirectional hidden state.
 \vspace{-0.5em}\item \textbf{Parallel decoding and selection.} We decode a set of length candidates in parallel at near single-candidate cost via a slot-wise attention mask, and select the final span with a post-hoc coherence score in one pass.
 \vspace{-0.5em}\item \textbf{Extensive validation.} Across five DLMs and eight code and text benchmarks, \ours surpasses the strongest baselines by $+4.8$ pass rate on code and $+6.0$ BLEU-2 on text while running $1.82\times$ faster.
\end{itemize}

\section{Related Work}
\noindent \textbf{Diffusion Language Models.} DLMs generate text by iteratively denoising masked or corrupted sequences. The idea originates from early works such as D3PM~\citep{d3pm} and Diffusion-LM~\citep{diffusion-lm}, and is further developed by Masked Diffusion Language Models (MDLMs), which improve the modeling objective, sampling, and scalability~\citep{shi2024simplified,sahoo2024simple}. 
Recent models scale DLMs to larger parameters, including research previews such as Gemini-Diffusion and Seed Diffusion~\citep{seed-diffusion} and open-sourced models such as LLaDA~\citep{llada} and Dream~\citep{dream}, reaching competitive performance across tasks including code generation, instruction following, and reasoning~\citep{dreamcoder,llada2.0}. 
Owing to their bidirectional attention, DLMs can condition on both left and right context, which makes them naturally suited to infilling~\citep{survey}.

\noindent \textbf{Infilling and Adaptive-Length Decoding.} Infilling has been extensively studied in the AR paradigm, including FIM-style training~\citep{humaneval-infilling} and code-specialized models~\citep{incoder,starcoder,codellama}, all of which decode the missing span left-to-right via causal attention. While effective, AR models can only access the suffix indirectly through prompt conditioning, motivating bidirectional alternatives such as DLMs that natively condition on both prefix and suffix~\citep{dream}. However, DLMs typically require the number of masked positions to be specified before decoding, restricting them to fixed-length generation~\citep{efficient-survey}. Recent studies address this by adaptive length decoding. DAEDAL~\citep{daedal} dynamically expands the generation canvas during inference, but only focuses on appending at the end of sequences. DreamOn~\citep{dreamon} introduces explicit length-changing operations into the diffusion process, but requires fine-tuning the DLM, and potentially compromises general-purpose capabilities. FlexMDM~\citep{flexmdm} and DDOT~\citep{ddot} also introduces additional fine-tuning cost. CAL~\citep{cal} searches for suitable lengths using early-step denoising confidence, but incurs extra forward passes and is highly sensitive to the initial length. 
In contrast, our approach uses a lightweight probe to directly predict the target length, and a multi-slot decoding scheme that generates and selects among multiple length candidates in parallel. This yields adaptive-length infilling with minimal additional cost and no reliance on a preset initial length.
\section{Preliminaries}

\label{sec:prelim}

\noindent 
\textbf{Diffusion Language Models.}
Let $x = (x_1, \dots, $ $x_M)$ be a token sequence over vocabulary $\mathcal{V}$, augmented with a mask token $\mathtt{[M]}$. DLMs define a forward process that independently replaces each $x_i$ with $\mathtt{[M]}$ with probability $t \in [0, 1]$, yielding a corrupted sequence $x^{(t)}$. A network $p_\theta(x^{(0)} \mid x^{(t)})$ is trained to reconstruct the original tokens at masked positions \citep{sahoo2024simple, shi2024simplified}. Unlike auto-regressive models, DLMs use bidirectional attention, so every prediction conditions on all other positions on both sides.
At inference, given a prefix $P$ and a target length $L$, decoding starts from $x^{(K)} = P \oplus \mathtt{[M]}^L$, where $\oplus$ denotes concatenation, and proceeds for $K$ steps. At each step $k = K, \dots, 1$, the model produces a distribution over $\mathcal{V}$ at every masked position; a subset of positions with the highest confidence are unmasked, while the rest remain masked for step $k{-}1$. The final $x^{(0)}$ contains no masks. A central limitation is that $L$ must be specified \emph{before} decoding and remains fixed throughout.

\noindent \textbf{Infilling Tasks for DLMs.} 
The inference setting above fills a masked region given only a prefix $P$. Infilling extends this setting by additionally fixing a suffix $S$ as a right-side condition, which DLMs can exploit directly thanks to their bidirectional attention and any-order generation. Given a prefix $P$ and a suffix $S$, the infilling task asks for a middle span $M = (w_1, \dots, w_{L^*})$ with $w_i \in \mathcal{V}$, such that the concatenation $P \oplus M \oplus S$ maximizes a task-specific quality metric $\mathcal{Q}$ (e.g., pass rate for code, BLEU-2 for text). The gold length $L^*$ is unknown at inference time, and $\mathcal{Q}$ is inaccessible during decoding.
DLMs instantiate the masked region as $P \oplus \mathtt{[M]}^{L_0} \oplus S$ at a preset length $L_0$ and run iterative masked decoding to fill it. 
\section{Method}

\begin{figure*}[!t]
\centering{
\includegraphics[scale=0.9]{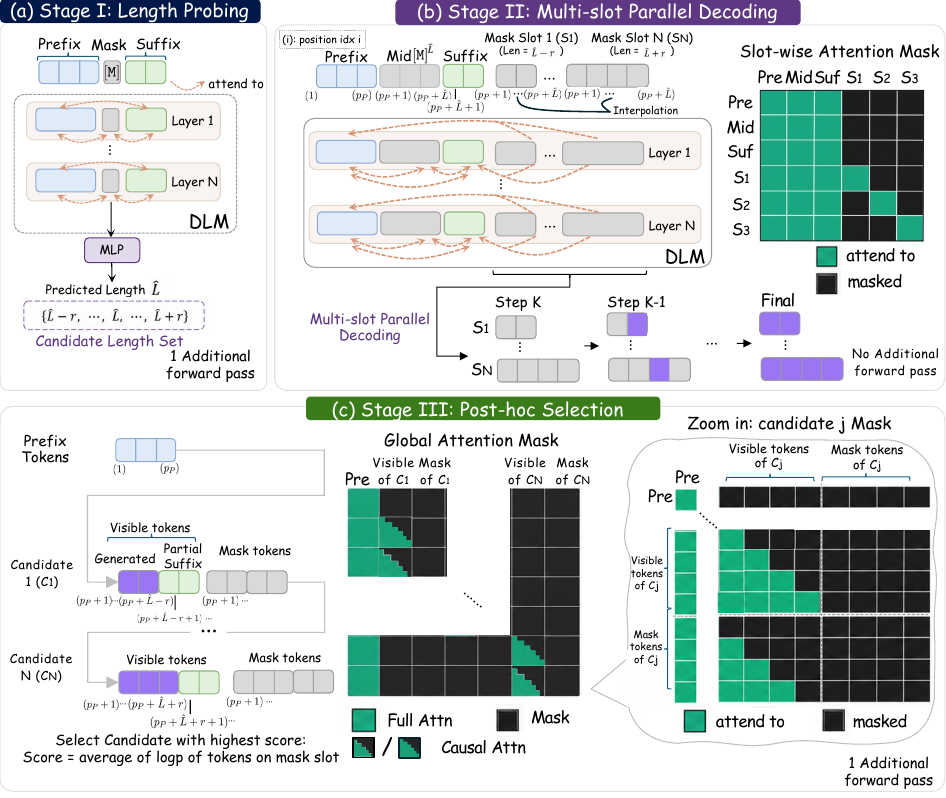}
\caption{\textbf{Pipeline of the proposed method.} (a) Stage I probes the hidden state of a mask token to predict the target length $\hat L$; (b) Stage II decodes the candidate lengths $\{\hat{L}-r,\dots,\hat{L}+r\}$ in parallel with a slot-wise attention mask; (c) Stage III selects the final span with post-hoc score in one forward pass via a well-designed attention mask.
\label{fig:method}
}\vspace{-0.7em}}
\end{figure*}

\subsection{Overview}
We cast the infilling task as two coupled subproblems: (1)~determining the length of the mask span and (2)~generating content coherent with both the prefix and the suffix. The two subproblems are entangled, since the appropriate length depends on what to generate, yet DLMs can only generate with fixed lengths. \ours decouples them through a three-stage design shown in Fig.~\ref{fig:method}. (1)~Stage I predicts a target length $\hat L$ directly from hidden states of mask token through a single forward pass with a lightweight MLP probe, avoiding iterative length search. Since the generation is sensitive to lengths, we expand $\hat L$ into a small candidate set $\{\hat L-r,\cdots,\hat L+r\}$ to enhance predictive reliability. (2)~Stage II generates all candidate lengths in parallel at nearly the cost of decoding one, where each candidate length occupies a separate decoding region (a slot) and a slot-wise attention mask isolates the slots so they share prefix/suffix context but do not leak into one another, accelerating generation. (3)~Stage III selects the final span from candidates via a post-hoc score that jointly measures each candidate's internal coherence and how naturally the suffix continues from it in a single additional forward pass, equipped with a well-designed attention mask mechanism. Throughout the pipeline, we only introduce two passes, the probing pass and the scoring pass. Also, by predicting the length directly rather than searching from a preset value, \ours requires no initial lengths, yielding consistent performance across datasets.

\subsection{Length Probing}
The first stage estimates the length of the missing span directly from hidden states, as shown in Fig.~\ref{fig:method}(a). Given a prefix $P$ and suffix $S$, we construct a \textit{probing sequence} by inserting a single mask token between them, denoted as $x_{\mathrm{probe}} = P \oplus \texttt{[M]} \oplus S$, and run a single forward pass of the frozen DLM backbone. We read the hidden state $h_{\mathtt{[M]}}\in\mathbb{R} ^d$ of the mask position at a certain layer. Because the backbone uses bidirectional attention, $h_{\mathtt{[M]}}$ aggregates information from both prefix and suffix, making it a natural summary of how much content the missing span is expected to hold. Rather than using only the mask hidden state, we form the probe representation
by concatenating the mean-pooled hidden states of the last four prefix tokens,
the mask hidden state, and the mean-pooled hidden states of the first four
suffix tokens
$
h =
[\mathrm{MeanPool}(h_{p-4:p-1});
h_{\texttt{[M]}};
\mathrm{MeanPool}(h_{s:s+3})]
\in \mathbb{R}^{3d}
$. This provides a richer representation than the single confidence scores used by DAEDAL~\citep{daedal} and CAL~\citep{cal}. Then, a lightweight 3-layer MLP probe $f_\phi(\cdot)$ maps $h$ to a scalar length estimate
\begin{equation}\hat L =  \big\lfloor \exp \big(f_\phi(h)\big)\big\rceil,\end{equation}
where $\lfloor \cdot\rceil$ rounds to the nearest positive integer.  We predict $\log L$ rather than $L$ directly, which guarantees a positive estimate and reduces the influence of long-tailed length distribution. More information about the probe can be found in Appendix~\ref{sec:details_length_probe}. 
The lightweight probe does not touch the backbone, preserving the DLM ability without finetuning the backbone. Also, we only introduce one single forward pass in this stage to avoid presetting an initial length, with minimal overhead. We also validate the probing by comparing predicted lengths against the oracle, indicating that our probe predicts lengths more accurately than the baseline while requiring no preset initial length (Sec.~\ref{sec:length_probing_effectiveness}).

\subsection{Multi-slot Parallel Decoding}
Since the DLM generation is highly sensitive to the length of the mask slot, we expand the length $\hat L$ obtained by the probe within a local radius $r$ into $\{\hat L-r,\cdots,\hat L+r\}$ to better reduce prediction errors, yielding $N=2r+1$ candidates, and assign the $j$-th candidate length $\ell_j$ to a separate decoding slot, which is a contiguous block of mask tokens that is decoded independently. But it yields much more decoding cost, therefore we propose a \textit{multi-slot parallel decoding} strategy at a cost close to decoding one slot alone (Fig.~\ref{fig:method}(b)).

\noindent \textbf{Sequence Layout.} Since all the candidates share the prefix and suffix, we pack the candidate slots together with a shared context region, $P\oplus \mathtt{[M]}^{\hat L}\oplus S \oplus [S_1]\cdots[S_N]$, where each slot $S_j$ is initialized with $\ell_j$ mask tokens. The prefix, middle mask slot and suffix form a shared context, and they attend to each other and are attended by every slot $S_j$. In addition, to avoid interference between mask slots $S_j$, we isolate the slots through a slot-wise attention mask. Each slot $S_j$ attends only to its own tokens and the shared context $\{P,\mathtt{[M]}^{\hat L}, S\}$, but not to any other mask slot $S_{j'}(j'\neq j)$. Because each slot sees nearly the same context as in single-slot generation, its denoising behavior, and thus its output, closely matches that of decoding the slot alone. This holds only when each slot is given appropriate position ids, which we address below.

\noindent \textbf{Position IDs.}
A subtlety is how to assign position ids to slots of different lengths. The key principle is that a DLM places each predicted token at the position dictated by its position id. If the mask span is given more position ids than the content needs, the model treats the extra positions as space still to be filled and under-generates.
We illustrate this with a concrete example, in which $\texttt{[TODO]}$ denotes the span to be filled.
\vspace{-0.5em}
\begin{center}
\begin{tabular}{@{}l@{}}
\pid{I}{1}\texttt{ have 13 books, and you have 23 books.}\\[-0.2em]
\texttt{We }\pid{have}{17}\texttt{ [TODO] books.}
\end{tabular}
\end{center}
\vspace{-0.5em}
The numbers below the text are the position ids assigned by LLaDA. The intended answer is "\texttt{36}" (two tokens). When the gap is given exactly two position ids, the model correctly produces "\texttt{36}". But when we keep a single mask token yet leave a one-position gap between it and one side of the context, the model fills only half the answer: leaving the gap on the suffix side makes the mask take position id "\texttt{(18)}" while the suffix resumes at "\texttt{(20)}", and the model outputs only "\texttt{3}" (the first token of "\texttt{36}"); leaving the gap on the prefix side instead yields "\texttt{6}" (the second token). In each case the model anchors its output to the position ids and leaves the gap unfilled, i.e., it under-generates.
It motivates a no-gap mask positional id design. The mask slot should start at the position id immediately after the prefix and end immediately before the suffix, leaving no gaps from them.

We therefore fix the suffix anchor and \emph{interpolate} each slot's position ids over a shared gap interval, so that all slots present an identical layout to the suffix regardless of their length. Let the last prefix token have position id $p_P$. We fix the suffix to start at $p_S = p_P + \hat L + 1$, leaving a gap interval $[\,p_P+1,\ p_S-1\,]$ of width $\hat L$. For a slot $S_j$ of length $\ell_j$, its $k$-th token ($k=1,\dots,\ell_j$) is assigned
\vspace{-0.5em}
\begin{equation}
\vspace{-0.5em}
\mathrm{pos}(S_j, k) = (p_P+1) + (k-1)\cdot \frac{\hat L - 1}{\ell_j - 1},
\label{eq:posid}
\end{equation}
linearly interpolating the $\ell_j$ tokens between $p_P+1$ and $p_S-1$. Every slot thus spans the same interval and the suffix is anchored at $p_S$ for all candidates, eliminating the length-dependent misalignment above. This design is critical: padding-based alternatives that left- or right-align the slot leave a position-id gap between the candidate and one side of the context, which leads to under-generation by the mechanism above. In contrast, interpolation removes this gap and substantially outperforms both padding variants (Sec~\ref{app:posid}).

\noindent \textbf{Parallel Decoding.} Within each slot, we run the original denoising/unmasking process. Since the mask slots do not interfere with each other and can be denoised in parallel, we obtain $N$ infill candidates $\{C_1, \cdots, C_N\}$ in almost the same time as decoding a single slot. This also saves KV-cache computation, since the shared context $\{P,\mathtt{[M]}^{\hat L}, S\}$ remains unchanged throughout decoding. The middle $\mathtt{[M]}^{\hat L}$
produces no output, and it merely anchors the shared position-id interval, and its KV is identical across all slots and steps.  This motivates placing the shared context first in the sequence layout ($P\oplus \mathtt{[M]}^{\hat L}\oplus S \oplus [S_1]\cdots[S_N]$), so that its \textit{KV can be cached and reused} across decoding steps. Please see Fig.~\ref{fig:method}(c) for an illustration.

\subsection{Post-hoc Selection}
\begin{figure}[h]
    \centering
    \includegraphics[scale=1]{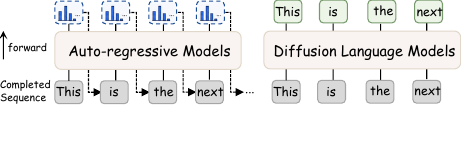}
    \vspace{-3.5em}
    \caption{Forward-pass targets in ARMs and DLMs.}
    \vspace{-0.5em}
    \label{fig:comparison}
\end{figure}

Stage~II returns $N$ infill candidates $\{C_1,\dots,C_N\}$ of different lengths,
and we must commit to one. AR models can score a sequence by its left-to-right next-token likelihood.
In contrast, DLMs are trained to predict the token \textit{at} a masked position rather than the \textit{next} token, so the distribution at an already-committed token is not a predictive distribution and cannot serve as a score (Fig.~\ref{fig:comparison}).
Our goal is to select the candidate that makes the \textit{completed} sequence
$P\oplus C_j\oplus S$ most coherent as a whole, rather than one that looks good in isolation. A natural score is the pseudo-log-likelihood, which masks each token in turn and reads its log-likelihood from the rest; however, this requires one forward pass per token and is prohibitively expensive. Instead, we fix the prefix as a shared condition across all candidates and score the inner part for internal coherence and the suffix for how naturally it continues from the candidate, all in a single forward pass.

\noindent \textbf{Masked Scoring Probes.} 
For candidate $C_j$ we lay out a visible block
$\boldsymbol{v}=C_j\oplus\tilde{S}=(v_1,\dots,v_{\ell_j+m})$, the $\ell_j$ committed candidate tokens
followed by the $m$ selected suffix tokens $\tilde{S}$-a small, representative subset of the suffix used to judge how well it continues from the candidate (introduced later)- and a block of \emph{scoring probes}, $\boldsymbol{\mu}=(\mu_1,\dots,\mu_{\ell_j+m})$, where each $\mu_k$ is a mask token and shares the position id of $v_k$.
We then carefully design an attention mask (Fig.~\ref{fig:method}(c)), enforcing a strictly causal read. The prefix $P$ is visible to the entire block.
(1)~Each \emph{visible} token
$v_k$ attends to $P$ and to $v_{\le k}$ (inclusive), so the visible
block stays internally causal and no visible representation contains information of a later token.
(2)~Each \emph{probe} $\mu_k$ attends to $P$ and to $v_{<k}$ only (strict), never to $v_k$ itself, to any later position, or to another probe, to prevent it from accessing $v_{\ge k}$.
Consequently the output at $\mu_k$ is $p_\theta(\cdot\mid P,v_{<k})$, and reading
it at the observed token gives $\log p_\theta(v_k\mid P,v_{<k})$ with no label
leakage; the first probe conditions on $P$ alone. The inclusive-causal visible
block is essential: were it bidirectional, $v_{<k}$ would already encode $v_k$ and
leak the token the probe is meant to predict. Since the suffix tokens follow the
full candidate in $v$, their probes automatically condition on all of $C_j$,
exactly the quantity needed to judge continuation.

\noindent \textbf{Selection Score.}
Let $c_k$ and $\tilde s_k$ be the $k$-th token in $C_j$ and $\tilde S_j$.
Partitioning the per-position log-probs of $v$ into its candidate and suffix
segments gives an inner score and a suffix score, 
\begin{align*}
s^{\text{in}}_j=\frac{1}{\ell_j}\sum_{k=1}^{\ell_j}\log p_\theta\!\left(c_k\mid P, c_{<k}\right),
\\
s^{\text{suf}}_j=\frac{1}{m}\sum_{t=1}^{m}\log p_\theta\!\left(\tilde{s}_t\mid P, C_j, {s}_{<t}\right),
\end{align*}
and we select the candidate maximizing their convex combination,
\begin{equation}
j^\star=\arg\max_{j}\;\alpha\,s^{\text{in}}_j+(1-\alpha)\,s^{\text{suf}}_j.
\label{eq:alpha}
\end{equation}
The inner score favors spans the model finds internally coherent, while the suffix
score favors spans after which the observed suffix becomes likely, penalizing
over- and under-generated spans that read fluently on their own but break the
transition into $S$. The weight $\alpha$ balances local fluency against
suffix alignment. We choose $\alpha=0.5$ in practice.

\noindent \textbf{Sparse Suffix Selection.}
Scoring the entire suffix is unnecessary, as the continuation signal concentrates
in a few tokens. We build $\tilde{S}$ from (i)~suffix head: the first few suffix tokens, which
carry the immediate transition, and (ii)~important tokens: the suffix tokens that received the
highest attention from the infilling region in Stage~II, which capture
longer-range dependence (e.g., some variables appear later in the code). These two groups need not be contiguous: the
intervening suffix tokens are not scored, but are kept as visible context so that each
scored token still conditions on the full true suffix preceding it, $\log p_\theta(\tilde{s}_t\mid P, C_j, s_{<t})$, with an attention mask.
We cap the considered suffix length at $32$ and use $m\in[4,8]$ (the two groups may overlap) scored tokens in practice.
For clarity, Fig.~\ref{fig:method}(c) depicts only the suffix head; the salient tokens
follow the same scoring structure and are omitted to save space.

\noindent \textbf{Single-pass Scoring.}
Finally, the $N$ candidate blocks are packed into one sequence and isolated by a
block-wise attention mask (Fig.~\ref{fig:method}(c), Global Attention
Mask): the visible tokens and probes of candidate $j$ attend only to the shared
prefix and to $C_j$'s own block, never across candidates, so the blocks cannot
interfere with each other. All candidates are therefore scored together in a single
additional forward pass.
\section{Experiments}

\begin{table*}[ht]
\begin{center}\setlength{\tabcolsep}{5pt}
\resizebox{1.0\linewidth}{!}{
\begin{tabular}{l cccccccc}
    \toprule
    \multirow[c]{2}{*}{\textbf{Datasets}}
    & \multicolumn{4}{c}{\textbf{Python}}
    & \multicolumn{2}{c}{\textbf{Other Language}}
    & \multicolumn{2}{c}{\textbf{Text}} \\
    \cmidrule(lr){2-5}\cmidrule(lr){6-7}\cmidrule(lr){8-9}
    & \textbf{Human-S} & \textbf{Human-M} & \textbf{MBPP-S} & \textbf{MBPP-M} & \textbf{Java} & \textbf{C/C++} & \textbf{Wiki} & \textbf{Arxiv} \\
    \midrule

    \multicolumn{9}{c}{\textbf{\textit{LLaDA-8B-Base}}}\\
    \midrule
    Backbone &  44.68& 15.92 & 37.45 &20.45  &34.09 & 34.84 & 19.90 / 34.48 & 13.23 / 30.30  \\
    DAEDAL & 44.99 &16.76 & 38.86 & 22.48 & 35.34 & 36.03& 10.82 / 20.78&6.65 / 16.61\\
    CAL & 64.74 & 24.61 & 54.49 & 32.21 & 50.31 & 49.22&23.00 / 37.25&15.69 / 32.69\\
    \ours & \textbf{71.35} & \textbf{28.20} & \textbf{57.49} & \textbf{38.19} &\textbf{57.10}&\textbf{54.57}&\textbf{29.29} / \textbf{44.96}&\textbf{20.28} / \textbf{39.82}\\
    \cdashline{1-9}
    \noalign{\vskip 2pt}
    \ours (Oracle) & 81.90 & 38.54 & 71.56 & 50.92 & 60.98&60.08&41.51 / 56.76&29.35 / 50.43\\
    \noalign{\vskip -1pt}
    \midrule

    \multicolumn{9}{c}{\textbf{\textit{LLaDA-8B-Instruct}}}\\
    \midrule
    Backbone & 49.73 & 17.62 & 37.47 & 21.15 & 32.42 & 33.40 & 18.07 / 32.77 & 12.57 / 29.44\\
    DAEDAL & 50.20 &18.68 & 39.17 & 23.02 &34.31 &34.81 &10.37 / 19.98 &6.56 / 16.64\\
    CAL & \textbf{69.77} & 26.93 &48.64 & 25.31 & 49.20 & 48.66 & 20.52 / 34.93 & 14.77 / 31.33 \\
    \ours & 67.86 &\textbf{27.38} &\textbf{52.27}& \textbf{36.86} & \textbf{52.66}& \textbf{52.35}& \textbf{25.58} / \textbf{41.91}&\textbf{18.94} / \textbf{38.13}  \\
    \cdashline{1-9}
    \noalign{\vskip 2pt}
    \ours (Oracle) & 78.99 & 41.03 & 72.38 & 56.67 &61.53&56.95 & 38.07 / 54.54 & 28.46 / 49.60
  \\
    \noalign{\vskip -1pt}
    \midrule

    \multicolumn{9}{c}{\textbf{\textit{LLaDA-MoE-Base}}}\\
    \midrule
    Backbone & 45.84 & 16.43 & 40.42 & 22.56&34.53&34.92&17.23 / 31.51&13.04 / 29.69 \\
    DAEDAL & 46.17 & 17.56 &41.50 & 24.77 & 37.25 & 37.14 & 8.90 / 18.48 & 6.06 / 15.70 \\
    CAL & 67.79 & 25.96 & 58.32& 34.45&53.06 & 50.41& 19.39 / 33.75&14.78 / 31.25\\
    \ours & \textbf{71.44} & \textbf{31.92}&\textbf{58.52}& \textbf{43.12} & \textbf{55.76} & \textbf{55.23}& \textbf{25.76} / \textbf{41.84}&\textbf{19.13} / \textbf{38.30}\\
    \cdashline{1-9}
    \noalign{\vskip 2pt}
    \ours (Oracle) & 81.80 & 41.34 & 67.56  &54.62 & 58.98 &58.77&37.51 / 53.41& 28.03 / 49.12\\
    \noalign{\vskip -1pt}
    \midrule

    \multicolumn{9}{c}{\textbf{\textit{Dream-7B-Base}}}\\
    \midrule
    Backbone & 42.79 & 16.13 & 40.12 & 22.15 & 33.90&34.30 & 19.12 / 33.75&13.56 / 30.97 \\
    DAEDAL &42.62 & 17.60& 41.22 &25.11 & 38.20& 37.88& 9.68 / 19.33 &6.72 / 17.22 \\
    CAL & 64.48 & 26.38 & 56.00 & 33.60 & 47.92 & 48.00 & 21.68 / 36.68&15.98 / 33.52\\
    
    \ours & \textbf{68.34} &\textbf{28.55} &\textbf{57.97}&\textbf{37.72} & \textbf{58.98}&\textbf{56.30}&\textbf{32.36} / \textbf{48.13} & \textbf{22.89} / \textbf{42.79}\\
    \cdashline{1-9}
    \noalign{\vskip 2pt}
    \ours (Oracle) &  79.77 & 39.19&70.91&52.11 &63.64 &62.63 &44.66 / 59.98 & 33.64 / 54.49\\
    \noalign{\vskip -1pt}
    \midrule

    \multicolumn{9}{c}{\textbf{\textit{Dream-Coder-7B-Base}}}\\
    \midrule
    Backbone & 49.64 & 19.00 & 43.94 & 25.62 & 36.89 & 37.66 & - & -\\
    DAEDAL& 50.85 & 21.14 & 47.79  & 30.80& 40.96 & 40.72 & - & -\\
    CAL & 68.56 & 28.24 & 61.35 & 38.91 & 51.14& 49.28 & - & -\\
    \ours & \textbf{71.35} & \textbf{32.14}&\textbf{62.59}&\textbf{47.28}& \textbf{59.87}&\textbf{59.01}&-&-\\
    
    \cdashline{1-9}
    
    \noalign{\vskip 2pt}
    \ours (Oracle) &  82.38 & 44.01 & 73.59  &63.41&62.20 & 62.88& - & - \\
    \noalign{\vskip -1pt}
    \bottomrule
\end{tabular}
}
\vspace{-0.6em}
\caption{Performance Comparison. We compare \ours with baselines on eight benchmarks across
five models.}
\label{tab:main}
\end{center}\vspace{-1.5em}
\end{table*}

\subsection{Experimental Setup}
\noindent \textbf{Models.} We apply \ours to five diffusion large language models (DLMs): LLaDA-8B-Base/Instruct~\citep{llada}, LLaDA-MoE-Base~\citep{lladamoe}, Dream-7B-Base~\citep{dream}, and Dream-Coder-7B-Base
~\citep{dreamcoder}. These models cover both general-purpose and code-oriented DLMs, spanning different model families (LLaDA/Dream), architectures (dense/MoE), and training recipes (base/instruction-tuned).

\noindent \textbf{Datasets.}
We evaluate on code and natural-language infilling benchmarks, all disjoint from probe training set. For code, we use HumanEval-Infilling~\citep{humaneval-infilling} with single-line (-S) and multi-line (-M) tasks, construct MBPP-S/-M from MBPP~\citep{mbpp} by a similar protocol, and build single-line Java and C/C++ tasks from MultiPL-E~\citep{multipl-e}. For text, we construct random contiguous word-span tasks from WikiText~\citep{wikitext} and arXiv abstracts. We report pass rate for code and BLEU-2/ROUGE-L for text. See Appendix~\ref{sec:datasets} for details.

\noindent\textbf{Baselines.}
We compare \ours with the backbone models (fixed-length decoding) and two dynamic decoding methods, DAEDAL~\citep{daedal} and CAL~\citep{cal}. Since all baselines require an initial generation length, we follow \citep{dreamon,cal} and report the average performance over $L \in \{4,8,16,32\}$. This provides a prior-free evaluation protocol, and per-length results are provided in Appendix. CAL is evaluated using its official implementation and hyperparameters. We adapt DAEDAL following the implementation in~\citep{cal} to fit the infilling setting. We also report comparison with DreamOn~\citep{dreamon} on text tasks. We also report results on \ours (Oracle), which uses the same generated candidate set as \ours but replaces
the Stage III selector with the ground-truth task evaluator, providing a upper bound on candidate selection.
All experiments run on a single NVIDIA A40 GPU.

\subsection{Main Results}

Table~\ref{tab:main} reports infilling quality across five DLMs and eight code and natural-language benchmarks. \ours delivers the strongest results on nearly every model-dataset combination, improving the most competitive baseline CAL by $+4.8$ pass rate on average across the code benchmarks and by $+6.0$ BLEU-2 points on text. These gains hold across model families (LLaDA/Dream), architectures (dense/MoE), training recipes (base/instruction-tuned), and both code and text, indicating that \ours fits different models and tasks.

\begin{figure*}[!th]
    \centering
    \includegraphics[width=0.95\linewidth]{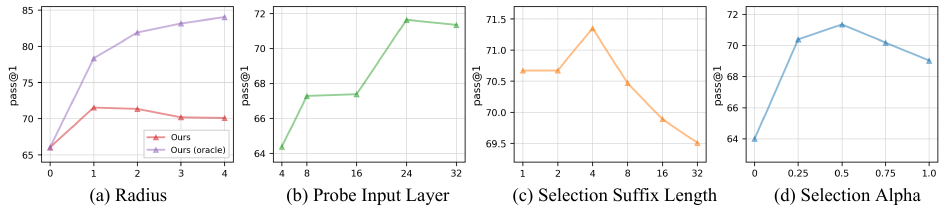}
    \vspace{-0.5em}
    \caption{Ablation studies on four design choices of \ours.}
    \label{fig:ablation}
    \vspace{-1em}
\end{figure*}

The baselines further illustrate why adaptive-length infilling is needed. DAEDAL, which only appends at the end of the canvas, cannot respect the suffix constraint and degrades even below the fixed-length backbone on text. CAL is the strongest baseline but still trails \ours while incurring repeated confidence-search passes and remaining sensitive to the preset initial length (Fig.~\ref{fig:teaser}(b)), whereas \ours needs no preset length at all. Finally, the oracle upper bound \ours (Oracle) shows that the candidate set almost always contains a strong completion, confirming that the multi-slot stage adequately covers the correct length. The remaining gap to the oracle indicates that the principal headroom lies in post-hoc selection rather than generation.

\subsection{Efficiency Analysis}
\label{sec:efficiency}
Beyond the WikiText results in Fig.~\ref{fig:teaser}(c), Table~\ref{tab:efficiency} compares quality and cost on HumanEval-S with LLaDA-8B-Base. \ours attains the highest Pass@1 while adding only 7\% wall-clock time over the fixed-length backbone, in contrast to CAL ($+94\%$) and DAEDAL ($+117\%$). Equivalently, \ours is $1.82\times$ faster than the strongest baseline CAL yet $6.6$ points more accurate. One of the reasons for this efficiency is that \ours introduces only two extra forward passes, the probing pass and the scoring pass, whereas the iterative length search in CAL and DAEDAL requires much more.

\begin{table}[H]
    \centering{
    \setlength{\aboverulesep}{0pt}
\setlength{\belowrulesep}{0pt}
\setlength{\extrarowheight}{2.5pt}
\setlength{\tabcolsep}{2pt}
\resizebox{0.9\columnwidth}{!}
{
    \begin{tabular}{l|c|cc}
     \toprule
\textbf{Method}   & \textbf{Pass@1} & \textbf{Extra Fwd. Passes} & \textbf{Time Cost}  \\
\midrule
Backbone &44.67& -  & 1.10 s \small{($+0\%$)} \\
DAEDAL    &44.80  &11.98& 2.29 s \small{(${\color{red!70}+117\%}$)} \\
CAL       &64.74  &13.79& 2.13 s \small{(${\color{red!70}+94\%}$)}\\
\midrule
\ours &\textbf{71.35} & 2.00& 1.18 s \small{(${\color{red!70}+7\%}$)} \\
\bottomrule
\end{tabular}
}
\vspace{-0.3em}
\caption{Efficiency of LLaDA-8B-Base on Human-S.}

    \vspace{-0.3em}
\label{tab:efficiency}
}
\end{table}

\subsection{Sensitivity to Design Choices}
\paragraph{Length probing and candidate set.} Fig.~\ref{fig:ablation}(b) varies the layer from which the probe reads the mask hidden state: Pass@1 increases monotonically with depth, consistent with deeper states aggregating richer bidirectional context from both prefix and suffix. We therefore read $h$ from the last layer for convenience. Fig.~\ref{fig:ablation}(a) varies the candidate radius $r$. The oracle upper bound keeps rising with $r$, confirming that a larger candidate set is more likely to contain the correct length. \ours, however, saturates beyond $r=2$, since candidate set is larger and the ratio of correct ones decreases. Also, we validate the effectiveness of length probing, as shown in Sec.~\ref{sec:length_probing_effectiveness}.

\paragraph{Post-hoc selection.} Fig.~\ref{fig:ablation}(c) varies the number of scored suffix tokens $m$. Pass@1 peaks at $m=4$
 and declines thereafter, supporting our suffix selection: the continuation signal concentrates in a few tokens near the transition, and scoring more of them only injects noise. Fig. ~\ref{fig:ablation}(d) varies the weight $\alpha$
between the inner and suffix scores. Performance peaks around $\alpha=0.5$ and drops at both extremes, showing that internal coherence and suffix continuation are complementary and neither alone suffices.

\subsection{Validation of Stages}
\paragraph{Length Probing Effectiveness (Stage I).}
\label{sec:length_probing_effectiveness}
We directly evaluate predicted lengths against the gold length $L^*$ on
Human-S with LLaDA-8B-Base, using MAE and Acc@$k$, the fraction of predictions
within $k$ tokens of $L^*$. For CAL, we sweep the preset initial length
$l\in\{4,8,16,32\}$, while \ours requires no initial length.

As shown in Table~\ref{tab:length_prediction}, CAL is highly sensitive to its
initialization: its MAE increases from 3.56 ($l{=}4$) to 24.13 ($l{=}32$),
while Acc@1 drops from 0.58 to 0.02. In contrast, \ours achieves the lowest
MAE (2.50) without tuning. Although CAL ($l{=}4$) has higher Acc@1, \ours
achieves substantially higher Acc@3 and Acc@5, which better reflects our
setting since \ours expands $\hat L$ into nearby candidate lengths and selects
among them. In particular, Acc@5 reaches 86.4\%, showing that the target
length is covered by the local candidate window in most cases.

\begin{table}[H]
    \centering
\resizebox{1.05\columnwidth}{!}{
    \begin{tabular}{lcccc}
        \toprule
        \textbf{Method} & \textbf{MAE} & \textbf{Acc@1} & \textbf{Acc@3} & \textbf{Acc@5} \\
        \midrule
        CAL ($l=4$)  & 3.5644  & 0.5808 & 0.7086 & 0.7909 \\
        CAL ($l=8$)  & 3.7106  & 0.5595 & 0.6912 & 0.7773 \\
        CAL ($l=16$) & 9.0600  & 0.2023 & 0.2730 & 0.3611 \\
        CAL ($l=32$) & 24.1336 & 0.0165 & 0.0203 & 0.0329 \\
        \ours       & 2.4974  & 0.4472 & 0.7822 & 0.8635 \\
        \bottomrule
    \end{tabular}
    }
    
    \vspace{-0.3em}
    \caption{Length prediction performance ($l$: initial length).}
    \label{tab:length_prediction}
    
    \vspace{-0.5em}
\end{table}

\paragraph{Effect of Slot Position-ID Design (Stage II).}
\label{app:posid}
We compare interpolation with left- and right-padding, which leave a position-ID gap on one side of the context. Since DLMs commonly use RoPE~\citep{su2024roformer}, interpolation at non-integer positions requires no additional tuning~\citep{chen2023extending,ddot}. For $\ell_j=1$, we set the position ID to $p_P+1$. As shown in Table~\ref{tab:posid}, interpolation substantially improves both Avg Pass@1 (51.6 vs.\ 28.2/31.6) and Oracle Pass@1 (81.9 vs.\ 73.2/74.2) over left/right padding. The gap persists under oracle selection, indicating that padding causes structural generation errors rather than merely poorer candidate selection.

We further compare against independent single-slot decoding. Given the same candidate lengths $\{\hat L-r,\ldots,\hat L+r\}$ produced by PILL, 
we decode each candidate separately in its own sequence, 
and oracle-select the best resulting candidate. Interpolation nearly matches this upper bound (81.9/51.6 vs.\ 83.6/52.9), showing that the shared multi-slot layout introduces little quality loss.

\begin{table}[t]
\centering
{\small
\begin{tabular}{lcc}
\toprule
\textbf{Position-ID Design} & \textbf{Oracle Pass@1} & \textbf{Avg Pass@1} \\
\midrule
Left-padding  & 73.2 & 28.2 \\
Right-padding & 74.2 & 31.6 \\
Interpolation (ours) & \textbf{81.9} & \textbf{51.6} \\
\midrule
Independent single-slot & 83.6 & 52.9 \\
\bottomrule
\end{tabular}
    \vspace{-0.3em}
\caption{Effect of slot position-id design (LLaDA-8B-Base, Human-S, $r{=}2$).}
\vspace{-1.5em}
\label{tab:posid}
}
\end{table}

\vspace{-0.3em}
\paragraph{Ablation of Post-hoc Selection Score (Stage III).}
\label{app:selection_ablation}
We ablate the post-hoc selection score in Table~\ref{tab:selection_ablation}.
Random uniformly selects a decoded candidate, while predicted-length only
uses the center candidate at $\hat{L}$ without post-hoc selection.
Inner-only and suffix-only correspond to $\alpha=1$ and $\alpha=0$ in
Eq.~\eqref{eq:alpha}, respectively, and our default combines both signals
with $\alpha=0.5$.
Random selection performs poorly despite the presence of strong candidates,
showing that effective post-hoc selection is necessary. Using the predicted
length alone is substantially stronger, while combining the inner and suffix
scores achieves the best performance, confirming that the two signals are
complementary.

\begin{table}[ht]
\centering

\small
\begin{tabular}{lc}
\toprule
\textbf{Selection Strategy} & \textbf{Pass@1} \\
\midrule
Random candidate & 51.62 \\
Predicted length only & 66.02 \\
Inner-only ($\alpha=1$) & 69.02 \\
Suffix-only ($\alpha=0$) & 63.99 \\
Inner + Suffix ($\alpha=0.5$) & \textbf{71.35} \\
\bottomrule
\end{tabular}

    \vspace{-0.3em}
\caption{
Ablation of the post-hoc selection score on LLaDA-8B-Base and HumanEval-S.
}\vspace{-1em}
\label{tab:selection_ablation}
\end{table}
\section{Conclusion}
We presented \ours, a backbone-frozen framework that turns fixed-length DLMs into adaptive-length infillers through three lightweight stages: probing the mask hidden state to predict the target length, decoding multiple length candidates in parallel via slot-wise attention, and selecting among them with a post-hoc coherence score. Adding only two forward passes and requiring no preset length, \ours outperforms prior adaptive-length baselines across five DLMs on both code and text infilling, running substantially faster than them.

\section*{Limitations}
While \ours is effective and efficient across a range of DLMs and infilling benchmarks, many practical scenarios involve complex editing
settings, such as nested spans and repository-level code patches, remain
interesting directions for future work. In addition, our experiments are conducted on DLMs up to roughly 7B/8B parameters on a single GPU. Whether the behavior of the length probe and the post-hoc selection scales to substantially larger diffusion backbones remains to be verified.

\section*{Acknowledgement}
\thanks{This work is supported by NSF (2324770). The content of the information in this document does not necessarily reflect the position or the policy of the Government, and no official endorsement should be inferred.  The U.S. Government is authorized to reproduce and distribute reprints for Government purposes notwithstanding any copyright notation here on.
}

\bibliography{custom}

@article{llada,
  title={Large language diffusion models},
  author={Nie, Shen and Zhu, Fengqi and You, Zebin and Zhang, Xiaolu and Ou, Jingyang and Hu, Jun and Zhou, Jun and Lin, Yankai and Wen, Ji-Rong and Li, Chongxuan},
  journal={Advances in Neural Information Processing Systems},
  volume={38},
  pages={50608--50646},
  year={2026}
}

@article{lladamoe,
  title={Llada-moe: A sparse moe diffusion language model},
  author={Zhu, Fengqi and You, Zebin and Xing, Yipeng and Huang, Zenan and Liu, Lin and Zhuang, Yihong and Lu, Guoshan and Wang, Kangyu and Wang, Xudong and Wei, Lanning and others},
  journal={arXiv preprint arXiv:2509.24389},
  year={2025}
}

@article{dream,
  title={Dream 7b: Diffusion large language models},
  author={Ye, Jiacheng and Xie, Zhihui and Zheng, Lin and Gao, Jiahui and Wu, Zirui and Jiang, Xin and Li, Zhenguo and Kong, Lingpeng},
  journal={arXiv preprint arXiv:2508.15487},
  year={2025}
}

@article{dreamcoder,
  title={Dream-coder 7b: An open diffusion language model for code},
  author={Xie, Zhihui and Ye, Jiacheng and Zheng, Lin and Gao, Jiahui and Dong, Jingwei and Wu, Zirui and Zhao, Xueliang and Gong, Shansan and Jiang, Xin and Li, Zhenguo and others},
  journal={arXiv preprint arXiv:2509.01142},
  year={2025}
}

@article{illada,
  title={Improved Large Language Diffusion Models},
  author={Nie, Shen and Min, Qiyang and Xu, Shaoxuan and Huang, Zihao and Song, Yuxuan and Shan, Yong and Lin, Yankai and Zhao, Wayne Xin and Li, Chongxuan and Wen, Ji-Rong},
  journal={arXiv preprint arXiv:2606.25331},
  year={2026}
}

@article{humaneval-infilling,
  title={Efficient training of language models to fill in the middle},
  author={Bavarian, Mohammad and Jun, Heewoo and Tezak, Nikolas and Schulman, John and McLeavey, Christine and Tworek, Jerry and Chen, Mark},
  journal={arXiv preprint arXiv:2207.14255},
  year={2022}
}

@article{mbpp,
  title={Program synthesis with large language models},
  author={Austin, Jacob and Odena, Augustus and Nye, Maxwell and Bosma, Maarten and Michalewski, Henryk and Dohan, David and Jiang, Ellen and Cai, Carrie and Terry, Michael and Le, Quoc and others},
  journal={arXiv preprint arXiv:2108.07732},
  year={2021}
}

@article{multipl-e,
  title={Multipl-e: A scalable and extensible approach to benchmarking neural code generation},
  author={Cassano, Federico and Gouwar, John and Nguyen, Daniel and Nguyen, Sydney and Phipps-Costin, Luna and Pinckney, Donald and Yee, Ming-Ho and Zi, Yangtian and Anderson, Carolyn Jane and Feldman, Molly Q and others},
  journal={arXiv preprint arXiv:2208.08227},
  year={2022}
}

@article{wikitext,
  title={Pointer sentinel mixture models},
  author={Merity, Stephen and Xiong, Caiming and Bradbury, James and Socher, Richard},
  journal={arXiv preprint arXiv:1609.07843},
  year={2016}
}

@article{daedal,
  title={Beyond fixed: Training-free variable-length denoising for diffusion large language models},
  author={Li, Jinsong and Dong, Xiaoyi and Zang, Yuhang and Cao, Yuhang and Wang, Jiaqi and Lin, Dahua},
  journal={arXiv preprint arXiv:2508.00819},
  year={2025}
}

@article{cal,
  title={Diffusion LMs Can Approximate Optimal Infilling Lengths Implicitly},
  author={Liu, Hengchang and Yang, Zhao and Su, Bing},
  journal={arXiv preprint arXiv:2602.00476},
  year={2026}
}

@article{dreamon,
  title={DreamOn: Diffusion Language Models For Code Infilling Beyond Fixed-size Canvas},
  author={Wu, Zirui and Zheng, Lin and Xie, Zhihui and Ye, Jiacheng and Gao, Jiahui and Gong, Shansan and Feng, Yansong and Li, Zhenguo and Bi, Wei and Zhou, Guorui and others},
  journal={arXiv preprint arXiv:2602.01326},
  year={2026}
}

@article{d3pm,
  title={Structured denoising diffusion models in discrete state-spaces},
  author={Austin, Jacob and Johnson, Daniel D and Ho, Jonathan and Tarlow, Daniel and Van Den Berg, Rianne},
  journal={Advances in neural information processing systems},
  volume={34},
  pages={17981--17993},
  year={2021}
}

@article{diffusion-lm,
  title={Diffusion-lm improves controllable text generation},
  author={Li, Xiang and Thickstun, John and Gulrajani, Ishaan and Liang, Percy S and Hashimoto, Tatsunori B},
  journal={Advances in neural information processing systems},
  volume={35},
  pages={4328--4343},
  year={2022}
}

@article{sahoo2024simple,
  title={Simple and effective masked diffusion language models},
  author={Sahoo, Subham S and Arriola, Marianne and Schiff, Yair and Gokaslan, Aaron and Marroquin, Edgar and Chiu, Justin T and Rush, Alexander and Kuleshov, Volodymyr},
  journal={Advances in Neural Information Processing Systems},
  volume={37},
  pages={130136--130184},
  year={2024}
}

@article{shi2024simplified,
  title={Simplified and generalized masked diffusion for discrete data},
  author={Shi, Jiaxin and Han, Kehang and Wang, Zhe and Doucet, Arnaud and Titsias, Michalis},
  journal={Advances in neural information processing systems},
  volume={37},
  pages={103131--103167},
  year={2024}
}

@article{seed-diffusion,
  title={Seed diffusion: A large-scale diffusion language model with high-speed inference},
  author={Song, Yuxuan and Zhang, Zheng and Luo, Cheng and Gao, Pengyang and Xia, Fan and Luo, Hao and Li, Zheng and Yang, Yuehang and Yu, Hongli and Qu, Xingwei and others},
  journal={arXiv preprint arXiv:2508.02193},
  year={2025}
}

@article{llada1.5,
  title={Llada 1.5: Variance-reduced preference optimization for large language diffusion models},
  author={Zhu, Fengqi and Wang, Rongzhen and Nie, Shen and Zhang, Xiaolu and Wu, Chunwei and Hu, Jun and Zhou, Jun and Chen, Jianfei and Lin, Yankai and Wen, Ji-Rong and others},
  journal={arXiv preprint arXiv:2505.19223},
  year={2025}
}

@article{llada2.0,
  title={Llada2. 0: Scaling up diffusion language models to 100b},
  author={Bie, Tiwei and Cao, Maosong and Chen, Kun and Du, Lun and Gong, Mingliang and Gong, Zhuochen and Gu, Yanmei and Hu, Jiaqi and Huang, Zenan and Lan, Zhenzhong and others},
  journal={arXiv preprint arXiv:2512.15745},
  year={2025}
}

@inproceedings{bei2026mem,
  title={Mem-gallery: Benchmarking multimodal long-term conversational memory for mllm agents},
  author={Bei, Yuanchen and Wei, Tianxin and Ning, Xuying and Zhao, Yanjun and Liu, Zhining and Lin, Xiao and Zhu, Yada and Hamann, Hendrik and He, Jingrui and Tong, Hanghang},
  booktitle={Proceedings of the 64th Annual Meeting of the Association for Computational Linguistics (Volume 1: Long Papers)},
  pages={40750--40784},
  year={2026}
}

@article{survey,
  title={A survey on diffusion language models},
  author={Li, Tianyi and Chen, Mingda and Guo, Bowei and Shen, Zhiqiang},
  journal={arXiv preprint arXiv:2508.10875},
  year={2025}
}

@article{incoder,
  title={Incoder: A generative model for code infilling and synthesis},
  author={Fried, Daniel and Aghajanyan, Armen and Lin, Jessy and Wang, Sida and Wallace, Eric and Shi, Freda and Zhong, Ruiqi and Yih, Wen-tau and Zettlemoyer, Luke and Lewis, Mike},
  journal={arXiv preprint arXiv:2204.05999},
  year={2022}
}

@article{starcoder,
  title={Starcoder: may the source be with you!},
  author={Li, Raymond and Allal, Loubna Ben and Zi, Yangtian and Muennighoff, Niklas and Kocetkov, Denis and Mou, Chenghao and Marone, Marc and Akiki, Christopher and Li, Jia and Chim, Jenny and others},
  journal={arXiv preprint arXiv:2305.06161},
  year={2023}
}

@article{codellama,
  title={Code llama: Open foundation models for code},
  author={Roziere, Baptiste and Gehring, Jonas and Gloeckle, Fabian and Sootla, Sten and Gat, Itai and Tan, Xiaoqing Ellen and Adi, Yossi and Liu, Jingyu and Sauvestre, Romain and Remez, Tal and others},
  journal={arXiv preprint arXiv:2308.12950},
  year={2023}
}

@article{flexmdm,
  title={Any-order flexible length masked diffusion},
  author={Kim, Jaeyeon and Cheuk-Kit, Lee and Domingo-Enrich, Carles and Du, Yilun and Kakade, Sham and Ngotiaoco, Timothy and Chen, Sitan and Albergo, Michael},
  journal={arXiv preprint arXiv:2509.01025},
  year={2025}
}

@article{hoogeboom2021argmax,
  title={Argmax flows and multinomial diffusion: Learning categorical distributions},
  author={Hoogeboom, Emiel and Nielsen, Didrik and Jaini, Priyank and Forr{\'e}, Patrick and Welling, Max},
  journal={Advances in neural information processing systems},
  volume={34},
  pages={12454--12465},
  year={2021}
}

@article{py150,
  title={Probabilistic model for code with decision trees},
  author={Raychev, Veselin and Bielik, Pavol and Vechev, Martin},
  journal={ACM SIGPLAN Notices},
  volume={51},
  number={10},
  pages={731--747},
  year={2016},
  publisher={ACM New York, NY, USA}
}

@article{codecontests,
  title={Competition-level code generation with alphacode},
  author={Li, Yujia and Choi, David and Chung, Junyoung and Kushman, Nate and Schrittwieser, Julian and Leblond, R{\'e}mi and Eccles, Tom and Keeling, James and Gimeno, Felix and Dal Lago, Agustin and others},
  journal={Science},
  volume={378},
  number={6624},
  pages={1092--1097},
  year={2022},
  publisher={American Association for the Advancement of Science}
}

@inproceedings{huang2025large,
  title={Large language model simulator for cold-start recommendation},
  author={Huang, Feiran and Bei, Yuanchen and Yang, Zhenghang and Jiang, Junyi and Chen, Hao and Shen, Qijie and Wang, Senzhang and Karray, Fakhri and Yu, Philip S},
  booktitle={Proceedings of the eighteenth ACM international conference on web search and data mining},
  pages={261--270},
  year={2025}
}

@article{wiki,
  title={Pointer sentinel mixture models},
  author={Merity, Stephen and Xiong, Caiming and Bradbury, James and Socher, Richard},
  journal={arXiv preprint arXiv:1609.07843},
  year={2016}
}

@article{c4,
  title={Exploring the limits of transfer learning with a unified text-to-text transformer},
  author={Raffel, Colin and Shazeer, Noam and Roberts, Adam and Lee, Katherine and Narang, Sharan and Matena, Michael and Zhou, Yanqi and Li, Wei and Liu, Peter J},
  journal={Journal of machine learning research},
  volume={21},
  number={140},
  pages={1--67},
  year={2020}
}

@article{su2024roformer,
  title={Roformer: Enhanced transformer with rotary position embedding},
  author={Su, Jianlin and Ahmed, Murtadha and Lu, Yu and Pan, Shengfeng and Bo, Wen and Liu, Yunfeng},
  journal={Neurocomputing},
  volume={568},
  pages={127063},
  year={2024},
  publisher={Elsevier}
}

@article{chen2023extending,
  title={Extending context window of large language models via positional interpolation},
  author={Chen, Shouyuan and Wong, Sherman and Chen, Liangjian and Tian, Yuandong},
  journal={arXiv preprint arXiv:2306.15595},
  year={2023}
}

@inproceedings{ddot,
  title={Flexible-length text infilling for discrete diffusion models},
  author={Zhang, Andrew and Sivakumar, Anushka and Tang, Chia-Wei and Thomas, Chris},
  booktitle={Proceedings of the 2025 Conference on Empirical Methods in Natural Language Processing},
  pages={31332--31347},
  year={2025}
}

@article{lin2024duquant,
  title={Duquant: Distributing outliers via dual transformation makes stronger quantized llms},
  author={Lin, Haokun and Xu, Haobo and Wu, Yichen and Cui, Jingzhi and Zhang, Yingtao and Mou, Linzhan and Song, Linqi and Sun, Zhenan and Wei, Ying},
  journal={Advances in Neural Information Processing Systems},
  volume={37},
  pages={87766--87800},
  year={2024}
}

@article{lin2026duquant++,
  title={DuQuant++: Fine-grained Rotation Enhances Microscaling FP4 Quantization},
  author={Lin, Haokun and Jia, Xinle and Xu, Haobo and Yao, Bingchen and Guo, Xianglong and Wu, Yichen and Lu, Zhichao and Wei, Ying and Zhang, Qingfu and Sun, Zhenan},
  journal={arXiv preprint arXiv:2604.17789},
  year={2026}
}

@article{zhang2025ta,
  title={Ta-vla: Elucidating the design space of torque-aware vision-language-action models},
  author={Zhang, Zongzheng and Xu, Haobo and Yang, Zhuo and Yue, Chenghao and Lin, Zehao and Gao, Huan-ang and Wang, Ziwei and Zhao, Hao},
  journal={arXiv preprint arXiv:2509.07962},
  year={2025}
}

@article{zhang2025robochemist,
  title={Robochemist: Long-horizon and safety-compliant robotic chemical experimentation},
  author={Zhang, Zongzheng and Yue, Chenghao and Xu, Haobo and Liao, Minwen and Qi, Xianglin and Gao, Huan-ang and Wang, Ziwei and Zhao, Hao},
  journal={arXiv preprint arXiv:2509.08820},
  year={2025}
}

@inproceedings{xu2024slog,
  title={Slog: An inductive spectral graph neural network beyond polynomial filter},
  author={Xu, Haobo and Yan, Yuchen and Wang, Dingsu and Xu, Zhe and Zeng, Zhichen and Abdelzaher, Tarek F and Han, Jiawei and Tong, Hanghang},
  booktitle={Forty-first International Conference on Machine Learning},
  year={2024}
}

@article{lin2025quantization,
  title={Quantization meets dllms: A systematic study of post-training quantization for diffusion llms},
  author={Lin, Haokun and Xu, Haobo and Wu, Yichen and Guo, Ziyu and Zhang, Renrui and Lu, Zhichao and Wei, Ying and Zhang, Qingfu and Sun, Zhenan},
  journal={arXiv preprint arXiv:2508.14896},
  year={2025}
}

@article{efficient-survey,
  title={Efficient diffusion language models: A comprehensive survey},
  author={Lin, Haokun and Jia, Xinle and Liu, Shaozhen and Xia, Shujun and Huang, Weitao and Xu, Haobo and Li, Junyang and Xiao, Yicheng and Xing, Xingrun and Guo, Ziyu and others},
  year={2026},
  publisher={Authorea}
}

@article{lin2026benchmarking,
  title={Benchmarking Trustworthiness of SLMs: Pre-trained vs. Compressed},
  author={Lin, Haokun and Zhu, Kaijie and Xu, Haobo and Wu, Yichen and Lu, Zhichao and Zhang, Qingfu and Sun, Zhenan},
  journal={arXiv preprint arXiv:2608.11981},
  year={2026}
}

@inproceedings{xu2026prune,
  title={Prune as you generate: Online rollout pruning for faster and better rlvr},
  author={Xu, Haobo and Chen, Sirui and Qiu, Ruizhong and Yan, Yuchen and Luo, Chen and Cheng, Monica Xiao and He, Jingrui and Tong, Hanghang},
  booktitle={Proceedings of the 64th Annual Meeting of the Association for Computational Linguistics (Volume 1: Long Papers)},
  pages={13876--13893},
  year={2026}
}

@inproceedings{zeng2025interformer,
  title={Interformer: Effective heterogeneous interaction learning for click-through rate prediction},
  author={Zeng, Zhichen and Liu, Xiaolong and Hang, Mengyue and Liu, Xiaoyi and Zhou, Qinghai and Yang, Chaofei and Liu, Yiqun and Ruan, Yichen and Chen, Laming and Chen, Yuxin and others},
  booktitle={Proceedings of the 34th ACM International Conference on Information and Knowledge Management},
  pages={6225--6233},
  year={2025}
}

@article{zeng2025hierarchical,
  title={Hierarchical lora moe for efficient ctr model scaling},
  author={Zeng, Zhichen and Hang, Mengyue and Liu, Xiaolong and Liu, Xiaoyi and Lin, Xiao and Qiu, Ruizhong and Wei, Tianxin and Liu, Zhining and Yuan, Siyang and Yang, Chaofei and others},
  journal={arXiv preprint arXiv:2510.10432},
  year={2025}
}

@inproceedings{zeng2023parrot,
  title={Parrot: Position-aware regularized optimal transport for network alignment},
  author={Zeng, Zhichen and Zhang, Si and Xia, Yinglong and Tong, Hanghang},
  booktitle={Proceedings of the ACM web conference 2023},
  pages={372--382},
  year={2023}
}

@inproceedings{zeng2024hierarchical,
  title={Hierarchical multi-marginal optimal transport for network alignment},
  author={Zeng, Zhichen and Du, Boxin and Zhang, Si and Xia, Yinglong and Liu, Zhining and Tong, Hanghang},
  booktitle={Proceedings of the AAAI Conference on Artificial Intelligence},
  volume={38},
  number={15},
  pages={16660--16668},
  year={2024}
}

@inproceedings{zeng2023generative,
  title={Generative graph dictionary learning},
  author={Zeng, Zhichen and Zhu, Ruike and Xia, Yinglong and Zeng, Hanqing and Tong, Hanghang},
  booktitle={International Conference on Machine Learning},
  pages={40749--40769},
  year={2023},
  organization={PMLR}
}

@inproceedings{zeng2026harnessing,
  title={Harnessing consistency for robust test-time llm ensemble},
  author={Zeng, Zhichen and Yu, Qi and Lin, Xiao and Qiu, Ruizhong and Ning, Xuying and Wei, Tianxin and Yan, Yuchen and He, Jingrui and Tong, Hanghang},
  booktitle={Findings of the Association for Computational Linguistics: EACL 2026},
  pages={3528--3545},
  year={2026}
}

@article{lin2025toklip,
  title={Toklip: Marry visual tokens to clip for multimodal comprehension and generation},
  author={Lin, Haokun and Wang, Teng and Ge, Yixiao and Ge, Yuying and Lu, Zhichao and Wei, Ying and Zhang, Qingfu and Sun, Zhenan and Shan, Ying},
  journal={arXiv preprint arXiv:2505.05422},
  year={2025}
}

@article{xia2025medrek,
  title={Medrek: Retrieval-based editing for medical llms with key-aware prompts},
  author={Xia, Shujun and Lin, Haokun and Wu, Yichen and Zhou, Yinan and Li, Zixuan and Wan, Zhongwei and Xing, Xingrun and Zheng, Yefeng and Li, Xiang and Shan, Caifeng and others},
  journal={arXiv preprint arXiv:2510.13500},
  year={2025}
}

@article{li2026iv,
  title={IV-CoT: Implicit Visual Chain-of-Thought for Structure-Aware Text-to-Image Generation},
  author={Li, Zixuan and Lin, Haokun and Xiao, Yicheng and Li, Zhiwei and Song, Xinyang and Zheng, Zelong and He, Yong and Yao, Heng and Ding, Ke and Yu, Chao and others},
  journal={arXiv preprint arXiv:2606.24849},
  year={2026}
}

@inproceedings{yang2026dapq,
  title={DapQ-DiT: Distribution-Aware Post-Training Quantization for Efficient Generative Tasks in Diffusion Transformers},
  author={Yang, Lianwei and Lin, Haokun and Wu, Yichen and Sun, Zhenan and Gu, Qingyi},
  booktitle={Proceedings of the 2026 International Conference on Multimedia Retrieval},
  pages={2371--2380},
  year={2026}
}

@inproceedings{xing2026efficientllm,
  title={EfficientLLM: Unified Pruning-Aware Pretraining for Auto-Designed Compact Language Models},
  author={Xing, Xingrun and Liu, Zheng and Xiao, Shitao and Gao, Boyan and Liang, Yiming and Lin, Haokun and Zeng, Xianlin and Li, Guoqi and Zhang, Jiajun},
  booktitle={Proceedings of the 64th Annual Meeting of the Association for Computational Linguistics (Volume 1: Long Papers)},
  pages={7813--7830},
  year={2026}
}

@article{yang2026reshape,
  title={Reshape and rotate: Adaptive weight reshaping and fine-grained rotation for ultra-low-bit diffusion transformers quantization},
  author={Yang, Lianwei and Lin, Haokun and Wu, Yichen and Shan, Caifeng and Sun, Zhenan and Gu, Qingyi},
  journal={Neurocomputing},
  pages={133830},
  year={2026},
  publisher={Elsevier}
}

@article{zeng2026subspace,
  title={Subspace alignment for vision-language model test-time adaptation},
  author={Zeng, Zhichen and Bao, Wenxuan and Lin, Xiao and Qiu, Ruizhong and Wei, Tianxin and Ning, Xuying and Yan, Yuchen and Luo, Chen and Cheng, Monica Xiao and He, Jingrui and others},
  journal={arXiv preprint arXiv:2601.08139},
  year={2026}
}

@article{liu2024intactkv,
  title={Intactkv: Improving large language model quantization by keeping pivot tokens intact},
  author={Liu, Ruikang and Bai, Haoli and Lin, Haokun and Li, Yuening and Gao, Han and Xu, Zhengzhuo and Hou, Lu and Yao, Jun and Yuan, Chun},
  journal={arXiv preprint arXiv:2403.01241},
  year={2024}
}

@inproceedings{yu2026planetalign,
  title={Planetalign: A comprehensive python library for benchmarking network alignment},
  author={Yu, Qi and Zeng, Zhichen and Yan, Yuchen and Liu, Zhining and Jing, Baoyu and Qiu, Ruizhong and Azad, Ariful and Tong, Hanghang},
  booktitle={International Conference on Learning Representations},
  volume={2026},
  pages={66087--66115},
  year={2026}
}

@article{chen2026dflash,
  title={Dflash: Block diffusion for flash speculative decoding},
  author={Chen, Jian and Liang, Yesheng and Liu, Zhijian},
  journal={arXiv preprint arXiv:2602.06036},
  year={2026}
}

@article{li2026eagle,
  title={Eagle-3: Scaling up inference acceleration of large language models via training-time test},
  author={Li, Yuhui and Wei, Fangyun and Zhang, Chao and Zhang, Hongyang},
  journal={Advances in Neural Information Processing Systems},
  volume={38},
  pages={136737--136756},
  year={2026}
}

@inproceedings{xiao2024efficient,
  title={Efficient streaming language models with attention sinks},
  author={Xiao, Guangxuan and Tian, Yuandong and Chen, Beidi and Han, Song and Lewis, Mike},
  booktitle={International Conference on Learning Representations},
  volume={2024},
  pages={21875--21895},
  year={2024}
}

@article{zhang2026quantvla,
  title={Quantvla: Scale-calibrated post-training quantization for vision-language-action models},
  author={Zhang, Jingxuan and Hsieh, Yunta and Wan, Zhongwei and Lin, Haokun and Wang, Xin and Wang, Ziqi and Lei, Yingtie and Zhang, Mi},
  journal={arXiv preprint arXiv:2602.20309},
  year={2026}
}

@article{zhang2023h2o,
  title={H2o: Heavy-hitter oracle for efficient generative inference of large language models},
  author={Zhang, Zhenyu and Sheng, Ying and Zhou, Tianyi and Chen, Tianlong and Zheng, Lianmin and Cai, Ruisi and Song, Zhao and Tian, Yuandong and R{\'e}, Christopher and Barrett, Clark and others},
  journal={Advances in neural information processing systems},
  volume={36},
  pages={34661--34710},
  year={2023}
}

@article{liu2024kivi,
  title={Kivi: A tuning-free asymmetric 2bit quantization for kv cache},
  author={Liu, Zirui and Yuan, Jiayi and Jin, Hongye and Zhong, Shaochen and Xu, Zhaozhuo and Braverman, Vladimir and Chen, Beidi and Hu, Xia},
  journal={arXiv preprint arXiv:2402.02750},
  year={2024}
}

@inproceedings{wang2018acekg,
  title={Acekg: A large-scale knowledge graph for academic data mining},
  author={Wang, Ruijie and Yan, Yuchen and Wang, Jialu and Jia, Yuting and Zhang, Ye and Zhang, Weinan and Wang, Xinbing},
  booktitle={Proceedings of the 27th ACM international conference on information and knowledge management},
  pages={1487--1490},
  year={2018}
}

@inproceedings{yan2021dynamic,
  title={Dynamic knowledge graph alignment},
  author={Yan, Yuchen and Liu, Lihui and Ban, Yikun and Jing, Baoyu and Tong, Hanghang},
  booktitle={Proceedings of the AAAI conference on artificial intelligence},
  volume={35},
  number={5},
  pages={4564--4572},
  year={2021}
}

@inproceedings{yan2021bright,
  title={Bright: A bridging algorithm for network alignment},
  author={Yan, Yuchen and Zhang, Si and Tong, Hanghang},
  booktitle={Proceedings of the web conference 2021},
  pages={3907--3917},
  year={2021}
}

@article{ban2021ee,
  title={Ee-net: Exploitation-exploration neural networks in contextual bandits},
  author={Ban, Yikun and Yan, Yuchen and Banerjee, Arindam and He, Jingrui},
  journal={arXiv preprint arXiv:2110.03177},
  year={2021}
}

@inproceedings{wang2023networked,
  title={Networked time series imputation via position-aware graph enhanced variational autoencoders},
  author={Wang, Dingsu and Yan, Yuchen and Qiu, Ruizhong and Zhu, Yada and Guan, Kaiyu and Margenot, Andrew and Tong, Hanghang},
  booktitle={Proceedings of the 29th ACM SIGKDD Conference on Knowledge Discovery and Data Mining},
  pages={2256--2268},
  year={2023}
}

@article{yan2023trainable,
  title={From trainable negative depth to edge heterophily in graphs},
  author={Yan, Yuchen and Chen, Yuzhong and Chen, Huiyuan and Xu, Minghua and Das, Mahashweta and Yang, Hao and Tong, Hanghang},
  journal={Advances in Neural Information Processing Systems},
  volume={36},
  pages={70162--70178},
  year={2023}
}

@article{yan2023reconciling,
  title={Reconciling competing sampling strategies of network embedding},
  author={Yan, Yuchen and Jing, Baoyu and Liu, Lihui and Wang, Ruijie and Li, Jinning and Abdelzaher, Tarek and Tong, Hanghang},
  journal={Advances in Neural Information Processing Systems},
  volume={36},
  pages={6844--6861},
  year={2023}
}

@inproceedings{jing2024sterling,
  title={Sterling: Synergistic representation learning on bipartite graphs},
  author={Jing, Baoyu and Yan, Yuchen and Ding, Kaize and Park, Chanyoung and Zhu, Yada and Liu, Huan and Tong, Hanghang},
  booktitle={Proceedings of the AAAI Conference on Artificial Intelligence},
  volume={38},
  number={12},
  pages={12976--12984},
  year={2024}
}

@inproceedings{yan2022dissecting,
  title={Dissecting cross-layer dependency inference on multi-layered inter-dependent networks},
  author={Yan, Yuchen and Zhou, Qinghai and Li, Jinning and Abdelzaher, Tarek and Tong, Hanghang},
  booktitle={Proceedings of the 31st ACM International Conference on Information \& Knowledge Management},
  pages={2341--2351},
  year={2022}
}

@inproceedings{du2021new,
  title={New frontiers of multi-network mining: Recent developments and future trend},
  author={Du, Boxin and Zhang, Si and Yan, Yuchen and Tong, Hanghang},
  booktitle={Proceedings of the 27th ACM SIGKDD Conference on Knowledge Discovery \& Data Mining},
  pages={4038--4039},
  year={2021}
}

@inproceedings{yan2024pacer,
  title={Pacer: Network embedding from positional to structural},
  author={Yan, Yuchen and Hu, Yongyi and Zhou, Qinghai and Liu, Lihui and Zeng, Zhichen and Chen, Yuzhong and Pan, Menghai and Chen, Huiyuan and Das, Mahashweta and Tong, Hanghang},
  booktitle={Proceedings of the ACM Web Conference 2024},
  pages={2485--2496},
  year={2024}
}

@article{ban2023neural,
  title={Neural exploitation and exploration of contextual bandits},
  author={Ban, Yikun and Yan, Yuchen and Banerjee, Arindam and He, Jingrui},
  journal={arXiv preprint arXiv:2305.03784},
  year={2023}
}

@inproceedings{roach2020canon,
  title={Canon: Complex analytics of network of networks for modeling adversarial activities},
  author={Roach, Shane and Ni, Connie and Kopylov, Alexei and Lu, Tsai-Ching and Xu, Jiejun and Zhang, Si and Du, Boxin and Zhou, Dawei and Wu, Jun and Liu, Lihui and others},
  booktitle={2020 IEEE International Conference on Big Data (Big Data)},
  pages={1634--1643},
  year={2020},
  organization={IEEE}
}

@inproceedings{yan2024topological,
  title={Topological anonymous walk embedding: A new structural node embedding approach},
  author={Yan, Yuchen and Hu, Yongyi and Zhou, Qinghai and Wu, Shurang and Wang, Dingsu and Tong, Hanghang},
  booktitle={Proceedings of the 33rd ACM International Conference on Information and Knowledge Management},
  pages={2796--2806},
  year={2024}
}

@article{yan2024thegcn,
  title={Thegcn: Temporal heterophilic graph convolutional network},
  author={Yan, Yuchen and Chen, Yuzhong and Chen, Huiyuan and Li, Xiaoting and Xu, Zhe and Zeng, Zhichen and Liu, Lihui and Liu, Zhining and Tong, Hanghang},
  journal={arXiv preprint arXiv:2412.16435},
  year={2024}
}

@inproceedings{li2024large,
  title={Large language model-guided disentangled belief representation learning on polarized social graphs},
  author={Li, Jinning and Han, Ruipeng and Sun, Chenkai and Sun, Dachun and Wang, Ruijie and Zeng, Jingying and Yan, Yuchen and Tong, Hanghang and Abdelzaher, Tarek},
  booktitle={2024 33rd International Conference on Computer Communications and Networks (ICCCN)},
  pages={1--9},
  year={2024},
  organization={IEEE}
}

@article{yang2024simce,
  title={SimCE: Simplifying Cross-Entropy Loss for Collaborative Filtering},
  author={Yang, Xiaodong and Chen, Huiyuan and Yan, Yuchen and Tang, Yuxin and Zhao, Yuying and Xu, Eric and Cai, Yiwei and Tong, Hanghang},
  journal={arXiv preprint arXiv:2406.16170},
  year={2024}
}

@inproceedings{yu2025joint,
  title={Joint optimal transport and embedding for network alignment},
  author={Yu, Qi and Zeng, Zhichen and Yan, Yuchen and Ying, Lei and Srikant, R and Tong, Hanghang},
  booktitle={Proceedings of the ACM on Web Conference 2025},
  pages={2064--2075},
  year={2025}
}

@article{zeng2025pave,
  title={Pave Your Own Path: Graph Gradual Domain Adaptation on Fused Gromov-Wasserstein Geodesics},
  author={Zeng, Zhichen and Qiu, Ruizhong and Bao, Wenxuan and Wei, Tianxin and Lin, Xiao and Yan, Yuchen and Abdelzaher, Tarek F and Han, Jiawei and Tong, Hanghang},
  journal={arXiv preprint arXiv:2505.12709},
  year={2025}
}

@inproceedings{yan2025answer,
  title={To Answer or Not to Answer (TAONA): A Robust Textual Graph Understanding and Question Answering Approach},
  author={Yan, Yuchen and Kolekar, Aakash and Genc, Sahika and Xu, Wenju and Huang, Edward W and Srinivasan, Anirudh and Jain, Mukesh and He, Qi and Tong, Hanghang},
  booktitle={Findings of the Association for Computational Linguistics: EMNLP 2025},
  pages={6360--6376},
  year={2025}
}

@inproceedings{liu2025few,
  title={Few-Shot Knowledge Graph Completion via Transfer Knowledge from Similar Tasks},
  author={Liu, Lihui and Wang, Zihao and Zhou, Dawei and Wang, Ruijie and Yan, Yuchen and Xiong, Bo and He, Sihong and Tong, Hanghang},
  booktitle={Proceedings of the 34th ACM International Conference on Information and Knowledge Management},
  pages={4960--4965},
  year={2025}
}

@article{lai2026minimax,
  title={Minimax sparse attention},
  author={Lai, Xunhao and Xu, Weiqi and Yang, Yufeng and Chen, Qiaorui and Xu, Yang and Zeng, Lunbin and Li, Xiaolong and Sun, Haohai and Zhu, Haichao and Zhang, Vito and others},
  journal={arXiv preprint arXiv:2606.13392},
  year={2026}
}

@article{zhu2025scaling,
  title={Scaling latent reasoning via looped language models},
  author={Zhu, Rui-Jie and Wang, Zixuan and Hua, Kai and Zhang, Tianyu and Li, Ziniu and Que, Haoran and Wei, Boyi and Wen, Zixin and Yin, Fan and Xing, He and others},
  journal={arXiv preprint arXiv:2510.25741},
  year={2025}
}

@article{cheng2026dspark,
  title={DSpark: Confidence-Scheduled Speculative Decoding with Semi-Autoregressive Generation},
  author={Cheng, Xin and Yu, Xingkai and Shao, Chenze and Li, Jiashi and Xiong, Yunfan and Qian, Yi and Zhu, Jiaqi and Ma, Shirong and Zhang, Xiaokang and Ye, Jiasheng and others},
  journal={arXiv preprint arXiv:2607.05147},
  year={2026}
}

@inproceedings{manvi2026zero,
  title={Zero-overhead introspection for adaptive test-time compute},
  author={Manvi, Rohin and Hong, Joey and Seyde, Tim and Labonne, Maxime and Lechner, Mathias and Levine, Sergey},
  booktitle={International Conference on Learning Representations},
  volume={2026},
  pages={109705--109723},
  year={2026}
}

@inproceedings{lin2024bemap,
  title={Bemap: Balanced message passing for fair graph neural network},
  author={Lin, Xiao and Kang, Jian and Cong, Weilin and Tong, Hanghang},
  booktitle={Learning on Graphs Conference},
  pages={37--1},
  year={2024},
  organization={PMLR}
}

@inproceedings{lin2026mixture,
  title={Mixture of Sequence: Theme-Aware Mixture-of-Experts for Long-Sequence Recommendation},
  author={Lin, Xiao and Tang, Zhicheng and Cong, Weilin and Hang, Mengyue and Wang, Kai and Wang, Yajuan and Zeng, Zhichen and Li, Ting-Wei and Yoo, Hyunsik and Liu, Zhining and others},
  booktitle={Proceedings of the ACM Web Conference 2026},
  pages={6469--6480},
  year={2026}
}

@article{lin2025moralise,
  title={Moralise: A structured benchmark for moral alignment in visual language models},
  author={Lin, Xiao and Liu, Zhining and Yang, Ze and Li, Gaotang and Qiu, Ruizhong and Wang, Shuke and Liu, Hui and Li, Haotian and Keswani, Sumit and Pardeshi, Vishwa and others},
  journal={arXiv preprint arXiv:2505.14728},
  year={2025}
}


\appendix
\section{Appendix}
\label{appendix}

\subsection{Details of the Length Probe}
\label{sec:details_length_probe}
\paragraph{Architecture.} The probe is implemented as a 3-layer MLP with hidden dimensions of 512 and 128. We use GeLU as the activation function.

\paragraph{Training Datasets.} For each model, we train a probe. We collect 196k training samples for each probe, from two sources: (1) a code corpus, including Py150~\citep{py150}, LeetCode, and CodeContests~\citep{codecontests}; and (2) a text corpus from C4~\citep{c4}. For each sample, we randomly mask a contiguous span from the original code or text sequence, using the remaining left and right contexts as the prefix and suffix, respectively. The length of the masked span is used as the target length. All training datasets are distinct from the evaluation datasets and are strictly non-overlapping. More details are provided in Sec.~\ref{sec:datasets}.

\paragraph{Data Construction.}
The probe is trained on auxiliary self-supervised infilling examples and does
not use any evaluation instances or test cases from evaluation datasets. Each example consists of a prefix, an original masked span, and a
suffix, with the token length of the masked span used as supervision.

For code, we construct examples from the public training splits of Py150,
LeetCode, and CodeContests. We retain contexts of at most 2,048 tokens under
the LLaDA tokenizer, mask non-comment code lines, and require non-empty prefix
and suffix contexts. For multilingual code, the target span is required to
contain at least four tokenizer tokens. We restrict Py150 examples to
top-level Python functions, extract LeetCode examples from its Python solution
field, and use only accepted submissions from CodeContests. No executable
tests from these auxiliary datasets are used.

For text, we sample examples from the English training split of C4. We
normalize each document, retain documents containing 80-400 words with
sufficient alphabetic content, exclude URL-heavy pages, and mask a random
contiguous span of 2-8 words while requiring at least 32 characters of both
prefix and suffix context. We apply no additional target-length filtering.
Since hidden representations differ across backbone models, we train a
separate probe for each backbone using the same data-construction pipeline.

\paragraph{Training Recipe.} For the length probe, we train a lightweight MLP regressor with a regression objective. Unless otherwise specified, the probe is optimized with AdamW using a learning rate of $1\times 10^{-3}$, weight decay of $1\times 10^{-4}$
, batch size of 16, and dropout rate of $0.1$. The model is trained for up to $50$ epochs with mean squared error (MSE) loss. 

\noindent
\textbf{Discussion.}
All adaptive-length methods incur some preparation cost; the key
difference lies in its \emph{form} and \emph{when it is paid}. The PILL
probe is trained only \emph{once} per backbone (about 100 minutes on a
single NVIDIA A40 GPU) and is thereafter reused across all tasks and
inputs with no further training, adding merely two forward passes at
inference time. In contrast, CAL is \emph{not} cost-free either: it
re-initializes its confidence parameters for \emph{each} setting (about
20 minutes, 2k samples), and this cost recurs \emph{online}, since its
iterative confidence search introduces many extra forward passes at
inference (Table~\ref{tab:efficiency}). DAEDAL likewise relies on
fitted/tuned components rather than operating fully out of the box.
PILL's cost is therefore a one-time, amortized expense, whereas the
baselines pay repeatedly at inference.
Moreover, the probe's training cost does \emph{not} stem from
overfitting to the evaluation distribution. As shown in our comparison
with DreamOn (Section~\ref{sec:dreamon}, Table~\ref{tab:dreamon}), the
text infilling tasks are out-of-distribution for our probe, which is
fitted only on code; PILL nonetheless outperforms DreamOn across all of
its length configurations on both datasets. This indicates that the
length prediction generalizes beyond its training domain, so the gains
arise from the method design rather than from in-distribution length
statistics.

\subsection{Per-Stage Overhead at a Fixed Length}
To attribute cost to each stage, we decompose \ours's runtime on
LLaDA-8B-Base (Human-S) against a controlled baseline: a single
fixed-length decoding at the \emph{same} predicted length $\hat{L}$
(denoted Basic Dec.). This isolates the cost added by each stage from any
effect of the decoding length itself, and therefore differs from the
end-to-end comparison in Table~\ref{tab:efficiency}, where the backbone is averaged over preset
lengths $L\in\{4,8,16,32\}$ and thus decodes at different lengths. As shown
in Table~\ref{tab:breakdown}, relative to this matched-length decoding, the
probe, multi-slot decoding, and selection add $8.9\%$, $3.4\%$, and $9.4\%$
respectively. Notably, the multi-slot stage adds almost nothing despite
generating all $2r+1$ candidates, confirming that the slot-wise parallel
design decodes the full candidate set at a cost close to a single slot;
the two lightweight probes account for the remainder.

\begin{table}[ht]
\centering{
\setlength{\tabcolsep}{2pt}
\resizebox{1.05\columnwidth}{!}
{
\begin{tabular}{lccccc}
     \toprule

\textbf{Breakdown }& \textbf{Basic Dec.}& \textbf{Probe} & \textbf{Multi-slot} & \textbf{Selection} & \textbf{Total} \\
\midrule
Time Cost& 0.967s & 0.086s& 0.033s & 0.091s & 1.177s\\
Overhead& - &8.9\% &3.4\% &9.4\% &21.7\%\\
\bottomrule
\end{tabular}
}
}
\caption{Per-stage overhead at a fixed length.}
\label{tab:breakdown}
\end{table}

\subsection{Full Results across Initial Lengths}
Table~\ref{tab:more_baselines} reports the per-length results that are
averaged into Table~\ref{tab:main} of the main paper, on three categories of tasks: Python (Human-S), other language (Java), and text (Wikitext).
Two patterns stand out. First, baseline quality is highly sensitive to the
preset length and varies non-monotonically with it: the backbone peaks at
$L{=}16$ on Human-S (56.24) but at $L{=}8$ on Wikitext (19.04), and even the
strongest baseline CAL drops from 70.86 at $L{=}4$ to 50.82 at $L{=}32$ on
Human-S, and from 22.69 to 5.35 over the same range on Wikitext. Crucially,
the best-performing length differs across datasets, so no single preset
length transfers, which is precisely the difficulty an adaptive method must
resolve. Second, larger lengths inflate the time cost monotonically while
quality does not follow, so the longer configurations pay more for worse
results. In contrast, \ours uses no preset length and a single run already
surpasses the best length-specific result of every baseline, at a time cost
close to the cheapest backbone configuration.

\begin{table*}[h]
    \centering
    
    {
    
\vspace{0.5em}
\textit{(a) Results on Human-S dataset}
\vspace{0.5em}

\resizebox{1.05\textwidth}{!}{
    \begin{tabular}{lcccccccccc} 
    \toprule 
    \multirow{2}{*}{\textbf{Init Length}} & \multicolumn{2}{c}{\textbf{4}} & \multicolumn{2}{c}{\textbf{8}} & \multicolumn{2}{c}{\textbf{16}} & \multicolumn{2}{c}{\textbf{32}} & \multicolumn{2}{c}{\textbf{Avg}} \\
    \cmidrule(lr){2-3} \cmidrule(lr){4-5} \cmidrule(lr){6-7} \cmidrule(lr){8-9} \cmidrule(lr){10-11}
    & \textbf{Pass@1} & \textbf{Time} & \textbf{Pass@1} & \textbf{Time} & \textbf{Pass@1} & \textbf{Time} & \textbf{Pass@1} & \textbf{Time} & \textbf{Pass@1} & \textbf{Time} \\
    \midrule
    Backbone & 22.36 & 0.279 & 51.21 & 0.567 & 56.24 & 1.155 & 48.89 & 2.411 & 44.68 & 1.103 \\
    DAEDAL & 29.72 & 0.519 & 53.34 & 1.123 & 53.63 & 3.095 & 43.27 & 4.417 & 44.99 & 2.288 \\
    CAL & 70.86 & 1.143 & 73.48 & 1.465 & 63.79 & 2.275 & 50.82 & 3.604 & 64.74 & 2.122 \\
    \ours & - & - & - & - & - & - & - & - & 71.35 & 1.176 \\
    \bottomrule 
\end{tabular}
}

\vspace{0.5em}
\textit{(b) Results on Java dataset}
\vspace{0.5em}

\resizebox{1.05\textwidth}{!}{
\begin{tabular}{lcccccccccc} 
    \toprule 
    \multirow{2}{*}{\textbf{Init Length}} & \multicolumn{2}{c}{\textbf{4}} & \multicolumn{2}{c}{\textbf{8}} & \multicolumn{2}{c}{\textbf{16}} & \multicolumn{2}{c}{\textbf{32}} & \multicolumn{2}{c}{\textbf{Avg}} \\
    \cmidrule(lr){2-3} \cmidrule(lr){4-5} \cmidrule(lr){6-7} \cmidrule(lr){8-9} \cmidrule(lr){10-11}
    & \textbf{Pass@1} & \textbf{Time} & \textbf{Pass@1} & \textbf{Time} & \textbf{Pass@1} & \textbf{Time} & \textbf{Pass@1} & \textbf{Time} & \textbf{Pass@1} & \textbf{Time} \\
    \midrule
    Backbone & 11.20 & 0.576 & 38.36 & 1.163 & 43.90 & 2.365 & 42.90 & 4.831 & 34.09 & 2.234 \\
    DAEDAL & 14.75 & 0.828 & 41.24 & 1.520 & 44.79 & 4.265 & 40.58 & 5.800 & 35.34 & 3.103 \\
    CAL & 49.78 & 2.379 & 54.99 & 3.039 & 50.89 & 4.432 & 45.57 & 7.038 & 50.31 & 4.222 \\
    \ours & - & - & - & - & - & - & - & - & 57.10&2.555\\
    \bottomrule 
\end{tabular}
}

\vspace{0.5em}
\textit{(c) Results on Wikitext dataset}
\vspace{0.5em}

\resizebox{1.05\textwidth}{!}{
\begin{tabular}{lcccccccccc} 
    \toprule 
    \multirow{2}{*}{\textbf{Init Length}} & \multicolumn{2}{c}{\textbf{4}} & \multicolumn{2}{c}{\textbf{8}} & \multicolumn{2}{c}{\textbf{16}} & \multicolumn{2}{c}{\textbf{32}} & \multicolumn{2}{c}{\textbf{Avg}} \\
    \cmidrule(lr){2-3} \cmidrule(lr){4-5} \cmidrule(lr){6-7} \cmidrule(lr){8-9} \cmidrule(lr){10-11}
    & \textbf{BLEU-2} & \textbf{Time} & \textbf{BLEU-2} & \textbf{Time} & \textbf{BLEU-2} & \textbf{Time} & \textbf{BLEU-2} & \textbf{Time} & \textbf{BLEU-2} & \textbf{Time} \\
    \midrule
    Backbone & 17.44 & 0.213 & 19.04 & 0.434 & 10.88 & 0.894 & 5.57 & 1.887 & 13.23 & 0.857 \\
    DAEDAL & 13.73 & 2.494 & 6.39 & 5.681 & 3.62 & 6.794 & 2.87 & 7.040 & 6.65 & 5.502 \\
    CAL & 19.27 & 0.553 & 22.69 & 0.883 & 15.44 & 1.513 & 5.35 & 2.387 & 15.69 & 1.334 \\
    \ours & - & - & - & - & - & - & - & - & 29.29 & 0.928\\
    \bottomrule 
\end{tabular}
    }}
    \caption{Baseline results under different initial lengths. }
    \label{tab:more_baselines}
\end{table*}

\subsection{More Results on Recent DLMs}
\label{app:illada}

To further evaluate the generalizability of \ours to stronger DLM backbones,
we apply it to iLLaDA-8B-Base~\citep{illada}, a recently proposed model that
improves upon LLaDA through scaled pre-training and enhanced fine-tuning.
We use the same \ours pipeline and evaluation protocol without modifying the
backbone. Table~\ref{tab:illada} reports the results on Java and C/C++.

\begin{table}[h]
\centering
\small
\begin{tabular}{lcccc}
\toprule
\textbf{Dataset} & \textbf{Backbone} & \textbf{DAEDAL} & \textbf{CAL} & \textbf{\ours} \\
\midrule
Java  & 35.95 & 38.75 & 52.99 & \textbf{57.43} \\
C/C++ & 37.90 & 40.39 & 52.94 & \textbf{55.88} \\
\bottomrule
\end{tabular}
\caption{Infilling performance on iLLaDA-8B-Base.}
\label{tab:illada}
\end{table}

\begin{table}[h]
\centering
\small
\begin{tabular}{lcccc}
\toprule
 & \textbf{Backbone} & \textbf{DAEDAL} & \textbf{CAL} & \textbf{\ours} \\
\midrule
Time / sample & 1.06s & 2.17s & 2.01s & \textbf{1.11s} \\
\bottomrule
\end{tabular}
\caption{Inference efficiency on iLLaDA-8B-Base under the same evaluation
protocol as Sec.~\ref{sec:efficiency}.}
\label{tab:illada_efficiency}
\end{table}

\ours consistently outperforms all baselines on both datasets while requiring
1.11s per sample, compared with 1.06s for fixed-length backbone decoding and
2.01s for CAL. Thus, \ours adds only about 5\% wall-clock overhead over the
backbone while retaining its efficiency advantage over iterative
adaptive-length decoding.

\subsection{Extension to Multi-span Infilling}
\label{app:multispan}

To evaluate \ours beyond single-span infilling, we construct a multi-span
infilling setting based on MBPP, where each instance contains multiple
disjoint missing spans. \ours handles all gaps jointly within a single
sequence: a single probing pass predicts the lengths of all missing spans,
multi-slot decoding generates their candidate spans in parallel, and the
resulting candidates are scored jointly.

\begin{table}[h]
\centering
\small
\begin{tabular}{lccc}
\toprule
\textbf{Method} & \textbf{Backbone} & \textbf{CAL} & \textbf{\ours} \\
\midrule
Pass Rate       & 33.80 & 40.73 & \textbf{45.59} \\
Time / sample   & 6.60s & 17.40s & \textbf{6.70s} \\
\bottomrule
\end{tabular}
\caption{Performance and efficiency on multi-span infilling constructed from
MBPP.}
\label{tab:multispan}
\end{table}

As shown in Table~\ref{tab:multispan}, \ours improves the pass rate by
11.79 points over fixed-length decoding and by 4.86 points over CAL.
Meanwhile, it incurs only about 1.5\% wall-clock overhead over the backbone
(6.70s vs.\ 6.60s) and is approximately 2.6$\times$ faster than CAL.
These results suggest that the probing and parallel-decoding design extends
naturally to multiple disjoint gaps without incurring iterative search for
each span.

\subsection{Robustness of Length Probing}
\label{app:probe_robustness}

\paragraph{Robustness to structural complexity.}
We first examine whether length probing remains reliable when the dependency
between the prefix and suffix becomes structurally more complex. To control
for span length, we restrict the analysis to Java and C/C++ examples with the
same gold infill length of 10 tokens. We parse each complete gold program and
measure structural complexity using (1) the maximum AST nesting depth within
the missing span and (2) the number of cross-boundary identifier reuses,
i.e., identifiers shared between the missing span and its prefix or suffix.
Examples are divided into low- and high-complexity groups at the median.

\begin{table}[h]
\centering

\setlength{\tabcolsep}{2pt}
\resizebox{1.05\columnwidth}{!}{
\small
\begin{tabular}{llcccc}
\toprule
\textbf{Dataset} & \textbf{Complexity} & \textbf{MAE} & \textbf{Exact} & \textbf{Within 1} & \textbf{Within 2} \\
\midrule
Java
    & Low  & 1.14 & 23.8 & 71.4 & 92.9 \\
    & High & 1.47 & 25.6 & 58.1 & 81.4 \\
C/C++
    & Low  & 1.59 & 21.7 & 58.7 & 78.3 \\
    & High & 1.91 & 21.3 & 46.8 & 70.2 \\
\bottomrule
\end{tabular}
}
\caption{Length-probing performance (\%) under different levels of structural
complexity. MAE is measured in tokens; the remaining columns report the
percentage of predictions that exactly match or fall within 1/2 tokens of
the gold length.}
\label{tab:probe_complexity}
\end{table}

As shown in Table~\ref{tab:probe_complexity}, greater structural complexity
moderately increases MAE and reduces near-exact prediction accuracy.
Nevertheless, 81.4\% and 70.2\% of high-complexity Java and C/C++ examples,
respectively, remain within two tokens of the gold length, matching the
default candidate radius $r=2$. This suggests that the probe remains
effective even in structurally complex, nested, and dependency-heavy
contexts.

\paragraph{Robustness to training data size.}
We further study the sensitivity of the length probe to the amount of
auxiliary training data. We train probes with different numbers of samples
and evaluate the resulting \ours models on Human-S with LLaDA-8B-Base.

\begin{table}[h]
\centering
\setlength{\tabcolsep}{2pt}
\resizebox{1.05\columnwidth}{!}{
\small
\begin{tabular}{lccccccc}
\toprule
\textbf{\# of Samples} & \textbf{1k} & \textbf{5k} & \textbf{10k} & \textbf{20k} & \textbf{50k} & \textbf{100k} & \textbf{CAL} \\
\midrule
Pass@1
& 59.24 & 66.21 & 67.96 & 69.31 & 70.86 & \textbf{71.35} & 64.74 \\
\bottomrule
\end{tabular}
}
\caption{Effect of the amount of auxiliary probe-training data on Human-S
with LLaDA-8B-Base.}
\label{tab:probe_data_size}
\end{table}

As shown in Table~\ref{tab:probe_data_size}, performance improves steadily
with more probe-training data but already surpasses CAL with 10k examples
(67.96 vs.\ 64.74). The improvement becomes modest beyond 50k samples,
suggesting that the probe does not critically depend on the full 100k
training set.

\subsection{Scaling of Multi-slot Parallel Decoding}
\label{app:multislot_scaling}

We study how the computational cost of multi-slot decoding scales with the
number of candidate lengths. We vary the candidate radius from $r=0$ to
$r=4$, corresponding to 1--9 candidates, and profile wall-clock time and
peak GPU memory on Human-S with LLaDA-8B-Base using a single NVIDIA A40 GPU.

\begin{table}[h]
\centering
\small
\resizebox{\columnwidth}{!}{
\begin{tabular}{lrrrrrr}
\toprule
\textbf{Radius $r$} & \textbf{0} & \textbf{1} & \textbf{2} & \textbf{3} & \textbf{4} & \textbf{Adaptive} \\
\midrule
\# Candidates
    & 1 & 3 & 5 & 7 & 9 & 3/5/7 \\
Time (s)
    & 0.82 & 0.95 & 1.10 & 1.27 & 1.44 & 1.13 \\
GPU Memory (GB)
    & 21.3 & 23.8 & 25.1 & 29.1 & 33.7 & 24.5 \\
Pass@1
    & 66.0 & 71.5 & 71.4 & 70.2 & 70.1 & 71.4 \\
Oracle Pass@1
    & 66.0 & 78.3 & 81.9 & 83.2 & 84.0 & 81.8 \\
\bottomrule
\end{tabular}}
\caption{Scaling of multi-slot decoding with the number of candidate lengths
on Human-S using LLaDA-8B-Base. The adaptive setting uses 3/5/7 candidates
according to prediction uncertainty.}
\label{tab:multislot_scaling}
\end{table}

As shown in Table~\ref{tab:multislot_scaling}, both runtime and memory increase approximately with the number of candidates,
since the shared prefix/suffix context is reused while only slot-specific
tokens grow with the candidate set. From one to nine candidates, runtime
increases from 0.82s to 1.44s (1.76$\times$) and memory from 21.3GB to
33.7GB (1.58$\times$). Meanwhile, Pass@1 saturates around $r=2$ even though
the oracle upper bound continues to increase, indicating that adding more
candidate lengths primarily improves coverage rather than final selected
performance. This motivates our default choice of $r=2$.

\subsection{Failure Modes}
\label{app:failure}

We analyze where the remaining errors of \ours arise by separating them into
probe coverage, candidate generation, and post-hoc selection failures.
Table~\ref{tab:failure_modes} summarizes the breakdown.

\begin{table}[h]
\centering
\small
\begin{tabular}{lc}
\toprule
\textbf{Failure Type} & \textbf{Ratio} \\
\midrule
Probe coverage failure & 24.3\% \\
\quad of which other candidates still pass & 10.2\% \\
Candidate generation failure & 3.6\% \\
Selection failure & 11.4\% \\
\bottomrule
\end{tabular}
\caption{Failure mode analysis of \ours on Human-S.}
\label{tab:failure_modes}
\end{table}

Probe coverage is the largest source of error. Interestingly, in 10.2\% of
cases associated with coverage failure, a nearby-length candidate still
produces a correct solution. Candidate generation failures are relatively
rare (3.6\%), while selection failures account for 11.4\%, consistent with
the remaining gap between \ours and \ours\,(Oracle).

\subsection{Potential of Uncertainty-aware Candidate Expansion}
\label{app:dynamic_radius}

Our default \ours uses a fixed candidate radius $r=2$ for all examples.
However, length predictions may have different levels of uncertainty,
suggesting that a fixed candidate set may not be optimal for every input.
We therefore first examine whether prediction uncertainty provides a useful
signal for dynamically adjusting the candidate radius.

\paragraph{Prediction uncertainty.}
We train five bootstrap length probes and use the standard deviation of their
predictions as an uncertainty estimate. Table~\ref{tab:probe_uncertainty}
reports the mean uncertainty grouped by gold span length.

\begin{table}[h]
\centering
\small
\begin{tabular}{lcccc}
\toprule
\textbf{Gold Length} & \textbf{1-8} & \textbf{9-16} & \textbf{17-32} & \textbf{$>$32} \\
\midrule
Mean Uncertainty & 0.106 & 0.123 & 0.161 & 0.185 \\
\bottomrule
\end{tabular}
\caption{Prediction uncertainty across different gold span lengths.}
\label{tab:probe_uncertainty}
\end{table}

Prediction uncertainty increases consistently with span length and has a
Pearson correlation of $0.32$ with the gold length. This suggests that longer
or more difficult spans tend to require a wider search neighborhood, while a
smaller candidate set may suffice for confident predictions.

\paragraph{Dynamic candidate radius.}
Motivated by this observation, we explore an uncertainty-aware candidate
expansion strategy. We divide examples into low-, medium-, and
high-uncertainty groups according to uncertainty terciles and assign 3, 5,
and 7 candidate lengths, respectively. Compared with the default fixed
radius $r=2$ (five candidates), the adaptive strategy improves Pass@1 from
71.35 to 71.44.

Although the improvement is modest on the current benchmark, this result
shows that prediction uncertainty can be used to dynamically allocate the
candidate budget without degrading performance. We view uncertainty-aware
candidate expansion as a promising direction for settings with more diverse
or longer span-length distributions.

\subsection{Additional Comparison Results}
\paragraph{Comparison with Oracle Best-Initialized CAL.}
\label{app:best_cal}
Our main evaluation follows a prior-free protocol that averages CAL over
initial lengths $L\in\{4,8,16,32\}$, since the optimal initialization is
unknown at deployment time and varies across datasets. To further examine
whether our results are driven by this averaging protocol, we compare \ours
against an oracle version of CAL that retrospectively selects its best
initial length on each test set.

\begin{table}[h]
\centering
\small
\begin{tabular}{lcc}
\toprule
\textbf{Dataset} & \textbf{CAL (Best Init.)} & \textbf{\ours} \\
\midrule
Human-S
& \textbf{73.48} ($L{=}8$)
& 71.35 \\

Human-M
& \textbf{31.61} ($L{=}32$)
& 28.20 \\

MBPP-S
& 56.93 ($L{=}8$)
& \textbf{57.49} \\

MBPP-M
& 37.79 ($L{=}16$)
& \textbf{38.19} \\

Java
& 55.10 ($L{=}8$)
& \textbf{57.10} \\

C/C++
& 54.32 ($L{=}8$)
& \textbf{54.57} \\
\bottomrule
\end{tabular}
\caption{Comparison with CAL using its best test-set initialization among
$L\in\{4,8,16,32\}$. This oracle initialization is selected retrospectively
and is not available in practical deployment.}
\label{tab:best_cal}
\end{table}

As shown in Table~\ref{tab:best_cal}, \ours outperforms oracle
best-initialized CAL on four of the six benchmarks, despite requiring no
preset initial length. Moreover, CAL's optimal initialization varies across
datasets, ranging from $L=8$ to $L=32$, illustrating the difficulty of
choosing a single transferable prior. On Human-S and Human-M, oracle CAL
performs better, but this comparison assumes access to test-set performance
when choosing the initialization. Thus, the results support our main
prior-free evaluation while providing a stronger upper-bound comparison for
CAL.

\paragraph{Comparison with DreamOn.}
\label{sec:dreamon}
DreamOn~\citep{dreamon} attains variable-length infilling by fine-tuning the DLM with explicit length-changing operations, whereas \ours requires no backbone-finetuning and leaves the backbone untouched. We compare on the text infilling tasks, which are out-of-distribution for both methods (Table~\ref{tab:dreamon}): DreamOn is trained on code, and for this comparison, our probe is also fitted only on code (with a separately trained code-only probe). The result is presented in Table~\ref{tab:dreamon}. 
In this out-of-domain setting, \ours outperforms DreamOn across all of its
length configurations (initial length $l\in\{1,4,8\}$) on both datasets,
showing that \ours transfers effectively across domains without backbone
fine-tuning. DreamOn's performance also varies noticeably with the initial
length, whereas \ours requires no preset length. We emphasize that DreamOn
operates under a different adaptation regime with task-specific backbone
fine-tuning, so this comparison is not intended to establish superiority over
its specialized fine-tuned system.

\begin{table}[h]
    \centering{
\resizebox{0.9\columnwidth}{!}
{
\setlength{\tabcolsep}{2pt}
    \begin{tabular}{lcccc}
         \toprule

\multirow{2}{*}{\textbf{Dataset}} & \multicolumn{2}{c}{\textbf{Wiki}} & \multicolumn{2}{c}{\textbf{Arxiv}} \\
\cmidrule(lr){2-3} \cmidrule(lr){4-5}
 & \textbf{BLEU-2} & \textbf{ROUGE-L} & \textbf{BLEU-2} & \textbf{ROUGE-L} \\
\midrule
DreamOn \footnotesize{($l$=1)}&18.72 & 33.43 & 13.76 & 30.25\\
DreamOn \footnotesize{($l$=4)}& 21.33 &36.09&15.62&33.17 \\
DreamOn \footnotesize{($l$=8)}& 17.39 & 31.34 & 14.50 & 31.57  \\
\ours & \textbf{23.67} & \textbf{38.89} & \textbf{16.11} & \textbf{34.43}\\
\bottomrule
\end{tabular}
}

    \vspace{-0.3em}
    \caption{Comparison with DreamOn ($l$: initial length). }
    \label{tab:dreamon}
}
\vspace{-1em}
\end{table}

\subsection{Quality-Efficiency Trade-off}
\begin{figure}
    \centering
    \includegraphics[scale=1.5]{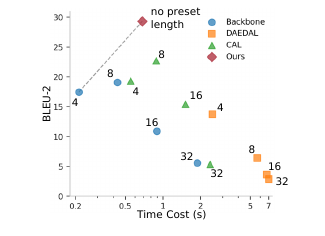}
    \caption{Quality-efficiency trade-off on Wikitext (BLEU-2 vs.\ wall-clock
    time, log-scaled $x$-axis). Each baseline is annotated with its preset
    length $L\in\{4,8,16,32\}$; \ours requires no preset length. \ours sits at
    the upper-left frontier and dominates all baseline configurations.}
    \label{fig:performance-time}
\end{figure}

Figure~\ref{fig:performance-time} visualizes the quality--time trade-off on
Wikitext, with each baseline shown at all four preset lengths
$L\in\{4,8,16,32\}$ and \ours shown as a single length-free point. \ours lies at
the upper-left frontier, attaining the highest BLEU-2 at a time cost lower
than nearly all baseline configurations; every baseline point is strictly
dominated. Sweeping a baseline along $L$ traces a curve that moves rightward
(more time) without consistently moving upward (better quality), illustrating
that tuning the preset length cannot recover the operating point \ours reaches
without any tuning.

\subsection{Additional Related Work}
\paragraph{Efficient Large Language Models.} With the rapid development of machine learning~\citep{xia2025medrek,ban2023neural,roach2020canon,yu2025joint,zeng2025pave} and foundation models~\citep{lin2025toklip,li2026iv}, recent studies have explored a wide range of applications, including recommendation~\citep{zeng2025hierarchical,zeng2025interformer,yan2022dissecting,yan2024thegcn,yang2024simce,li2024large,jing2024sterling,liu2025few,lin2026mixture}, graph learning~\citep{wang2018acekg,wang2023networked,du2021new,xu2024slog,yu2026planetalign,yan2021dynamic,yan2021bright,lin2024bemap}, reasoning~\citep{zeng2026harnessing,zeng2026subspace,yan2023trainable,yan2023reconciling}, and multi-modality tasks~\citep{zhang2025robochemist,zhang2025ta,bei2026mem,lin2025moralise}. Meanwhile, as foundation models continue to scale, their increasing computational and memory costs have motivated extensive research on efficient large language models. Existing studies improve efficiency from different perspectives, including small language model learning~\citep{xing2026efficientllm,lin2026benchmarking,yan2025answer}, model compression and quantization~\citep{lin2024duquant,lin2026duquant++,lin2025quantization,yang2026dapq,yang2026reshape,zhang2026quantvla}, efficient attention and long-context modeling~\citep{xiao2024efficient,zhang2023h2o}, KV-cache optimization~\citep{liu2024kivi,liu2024intactkv}, and efficient model architectures~\citep{lai2026minimax, zhu2025scaling}. These approaches aim to reduce the computational and memory overhead of large language models while preserving their modeling capabilities. One can refer to~\citep{efficient-survey}.

\paragraph{Efficient Decoding and Inference.} Recent advances in artificial intelligence~\citep{zeng2023generative,zeng2024hierarchical,yan2024pacer,yan2024topological,zeng2023parrot,ban2021ee} have led to increasingly capable but computationally intensive generative models~\cite{huang2025large}, making efficient inference an important research problem. Existing studies have explored accelerated decoding~\citep{xu2026prune}, parallel decoding~\citep{cheng2026dspark}, speculative decoding~\citep{chen2026dflash,li2026eagle}, and adaptive inference strategies~\citep{manvi2026zero} to reduce sequential computation and redundant model evaluations. Such efficiency is particularly important for diffusion language models, which generate sequences through iterative denoising and may require repeated forward passes during inference. In adaptive-length infilling, existing approaches further introduce additional computation to determine or adjust the generation length. In contrast, PILL directly predicts the target length, decodes multiple nearby length candidates in parallel, and selects the final candidate with a single additional scoring pass, enabling adaptive-length infilling with minimal inference overhead.

\subsection{Details of Datasets}
\label{sec:datasets}

All datasets and models used in this work are publicly available under licenses permitting research use, and we use them accordingly. Also, all evaluation results for test dataset are from a single run with deterministic decoding. Following all prior adaptive-length infilling methods we compare
against~\citep{daedal,cal,dreamon}, we report
results from a single run with deterministic decoding, which keeps the
comparison protocol identical across methods. 

Our use is consistent with these resources' intended research use, and the infilling benchmarks we derive from them are likewise intended for research only.

\paragraph{HumanEval.}
HumanEval-Infilling~\citep{humaneval-infilling} is an infilling benchmark derived from the original HumanEval code generation benchmark. It evaluates whether a model can complete missing code spans given both the left and right contexts. Following prior work, the benchmark contains several infilling settings, including single-line, multi-line, random-span, and random-span-light, which cover different granularities and locations of missing code.

\paragraph{MBPP.}
Mostly Basic Python Problems (MBPP)~\citep{mbpp} is a Python program synthesis benchmark consisting of 974 crowd-sourced programming problems. Each example contains a natural language task description, a reference Python solution, and test cases for functional correctness. The problems are designed to be solvable by entry-level programmers and mainly cover basic programming concepts, standard library usage, and simple algorithmic reasoning.

\paragraph{Py150.}
Py150~\citep{py150} is a large-scale Python corpus collected from GitHub repositories. The dataset contains 150k Python files, split into 100k training files and 50k evaluation files. The original release provides parsed Python ASTs as well as source files, and applies filtering steps such as removing duplicate files, excluding repository forks, keeping files that can be parsed, and focusing on permissive licenses. We use Py150 as a source of natural Python code for constructing length-probe training samples.

\paragraph{LeetCode.}
The LeetCode dataset\footnote{https://huggingface.co/datasets/greengerong/leetcode.} contains algorithmic programming problems from LeetCode, including problem titles, difficulty levels, descriptions, and reference solutions in multiple programming languages such as Python, Java, C++, and JavaScript. Compared with MBPP, these examples are generally more algorithmic and closer to interview-style coding tasks.

\paragraph{CodeContests.}
CodeContests~\citep{codecontests} is a competitive programming dataset released by DeepMind and used in the development of AlphaCode. It contains programming problems collected from multiple online judge platforms, such as Aizu, AtCoder, CodeChef, Codeforces, and HackerEarth. Each problem includes natural language statements, input-output test cases, and both correct and incorrect human submissions in multiple programming languages. We use it as a source of challenging algorithmic code.

\paragraph{MultiPL-E.}
MultiPL-E~\citep{multipl-e} is a multilingual code generation benchmark that translates unit-test-driven Python benchmarks, including HumanEval and MBPP, into 18 additional programming languages. It is designed to evaluate whether code generation models generalize beyond Python and across diverse programming languages. In our setting, MultiPL-E provides multilingual code examples and evaluation-style prompts for constructing infilling instances.

\paragraph{WikiText.}
WikiText~\citep{wikitext} is a language modeling dataset extracted from verified Good and Featured articles on Wikipedia. Compared with heavily preprocessed corpora such as Penn Treebank, WikiText preserves original casing, punctuation, numbers, and long-form article structure, making it suitable for evaluating long-range dependencies in natural language modeling. We use WikiText as a source of high-quality natural text for non-code infilling samples.

\paragraph{C4.}
The Colossal Clean Crawled Corpus (C4)~\citep{c4} is a large-scale cleaned web corpus derived from Common Crawl and introduced with the T5 framework. It contains diverse English web text after a series of filtering and cleaning steps. We use C4 as a broad-domain natural language corpus to improve the coverage of text-side length-probe training examples.

\paragraph{arXiv.}
The arXiv dataset\footnote{https://huggingface.co/datasets/UniverseTBD/arxiv-abstracts-large.} provides metadata and abstracts for scholarly articles across domains such as physics, computer science, statistics, electrical engineering, quantitative biology, and economics. We use arXiv abstracts as scientific-domain natural text, which complements the general-domain text from C4 and WikiText.

\subsection{Potential Risks}
\ours is a backbone-frozen method for diffusion language models: it does not modify the underlying backbone, and its lightweight probe is trained only on public corpora. We do not expect any personally identifying information in these resources. As our method builds on pretrained DLMs, it may inherit biases or factual errors present in these backbones, and the infilled spans it produces should be reviewed before use in sensitive applications.

All datasets used in our experiments are commonly used, publicly available
research datasets. We rely on their released and curated versions rather than
collecting data directly from individuals, and we do not use or intentionally
retain personally identifying information. We additionally inspected the
data-construction pipeline and did not identify fields intended to contain
personal identifiers or offensive content.
 
\subsection{Declaration of AI Assistance} 
We used AI tools for language polishing and part of the experimental implementation. These tools did not contribute to the idea, analysis, results, or scientific claims.

\end{document}